\documentclass{article}
\usepackage{ijcai25}

\usepackage{times}
\usepackage{soul}
\usepackage{url}
\usepackage[hidelinks]{hyperref}
\usepackage[utf8]{inputenc}
\usepackage[small]{caption}
\usepackage{graphicx}
\usepackage{amsmath}
\usepackage{amsthm} 
\usepackage{amssymb} 
\usepackage{booktabs}
\usepackage{algorithm}
\usepackage{algorithmic}
\usepackage[switch]{lineno}

\usepackage{float}
\usepackage{caption, subcaption, amsmath, amssymb,amsthm}
\usepackage{multicol}
\usepackage{xcolor} 
\usepackage{booktabs} 
\title{THEMIS: Unlocking Pretrained Knowledge with Foundation Model Embeddings for Anomaly Detection in Time Series} 

\author{
    Yadav Mahesh Lorik $^1$ \and Kaushik Sarveswaran $^2$ \and Nagaraj Sundaramahalingam $^3$ \and Aravindakumar Venugopalan $^4$ 
    \affiliations
    $^{1,2,3,4}$ Comcast India Engineering Center  
    \emails
    mahesh\_yadav@comcast.com $^1$, kaushik\_s@comcast.com $^2$
}

\begin{document}

\maketitle

\begin{abstract}
Time series anomaly detection forms a very crucial area in several domains but poses substantial challenges. Due to time series data possessing seasonality, trends, noise, and evolving patterns (concept drift), it becomes very difficult to set a general notion of what constitutes normal behavior. Anomalies themselves could be varied, ranging from a single outlier to contextual or collective anomalies, and are normally very rare; hence, the dataset is largely imbalanced. Additional layers of complexities arise due to the problems of increased dimensionality of modern time series, real-time detection criteria, setting up appropriate detection thresholds, and arriving at results that are interpretable. To embrace these multifaceted challenges, very strong, flexible, and interpretable approaches are required. This paper presents THEMIS, a new framework for time series anomaly detection that exploits pretrained knowledge from foundation models. THEMIS extracts embeddings from the encoder of the Chronos time series foundation model and applies outlier detection techniques like Local Outlier Factor and Spectral Decomposition on the self-similarity matrix, to spot anomalies in the data. Our experiments show that this modular method achieves SOTA results on the MSL dataset and performs quite competitively on the SMAP and SWAT$^*$ datasets.  Notably, THEMIS exceeds models trained specifically for anomaly detection, presenting hyperparameter robustness and interpretability by default. This paper advocates for pretrained representations from foundation models for performing efficient and adaptable anomaly detection for time series data.
\end{abstract}
\section{Introduction}

Time series data are generated in a plethora of fields, ranging from finance (fraud detection) to Industrial IoT (predictive maintenance) to healthcare (patient monitoring). This suggests that the need for robust and effective anomaly detection techniques is becoming critical and remains ever suspected \cite{Chandola2009}. Usually, an anomaly stands for the data points, sequences, or patterns that greatly diverge from normal behavior, and are thought to be significant \cite{Ahmed2016}. Anomaly detection (AD) in time series remains challenging due to factors like the scarcity, sparsity, and ambiguity of anomaly labels, often making supervised methods impractical where anomalies are rare and context-dependent. While some traditional approaches analyze residuals from forecasts \cite{Malhotra2015,Hyndman2018}, a broader challenge is the sensitivity of many AD techniques to parameter choices, necessitating laborious tuning and considerable domain knowledge \cite{Ren2019}. This sensitivity hinders scalability, adaptability, and robustness when faced with an increase either in the volume or in the heterogeneity of data \cite{BlazquezGarcia2021}. This motivates the need for approaches that are both effective and practical with minimal tuning overhead.

The recent paradigm-shifting developments in self-supervised learning have led to the creation of highly capable time series foundation models (TSFMs) such as Chronos \cite{Ansari2024}. These models are pretrained on an immense and varied corpus of time series data, which enables them to learn rich, nuanced, and generalizable temporal representations. Forecasting tasks, being inherently self-supervised with abundant training signals via next-step prediction, have been a primary focus for these FMs across diverse domains \cite{Ansari2024,gpt4ts}. Our premise is that the internal embeddings produced by these TSFMs can provide an effective representation of the central attributes and underlying dynamics of time series segments. In this regard, it would be reasonable to expect that, while normal behavior segments should form tight clusters in this embedding space, anomalous segments corresponding to a significant deviation should either occupy sparse regions or form separate, smaller clusters \cite{yue2022ts2vec}.

However, while TSFMs like Chronos \cite{Ansari2024} perform excellent forecasting, a direct approach to anomaly detection involves taking forecasts from these models, using a metric like Mean Squared Error (MSE) to determine the forecast error, and applying thresholds on them to label anomalies has inherent limitations. Several studies in the literature have argued that there are inherent limitations with direct repurposing of TSFMs in the way described for anomaly-related tasks. \cite{Shyalika2024TimeSF} argues that TSFMs are good forecasters but "...are limited in anomaly detection and prediction, with traditional statistical, as well as specialized deep learning models frequently outperforming them in AD performance, cost, and practicality." Their conclusion arises from the fact that such search is restricted by "the black-box nature of TSFMs and the lack of specialized designs for anomaly-related tasks" \cite{Shyalika2024TimeSF}. This highlights a gap in making use of the potential in TSFM representations for specialized tasks like Anomaly Detection.

\begin{figure*}[t] 
\centering
\includegraphics[width=0.6\textwidth, keepaspectratio]{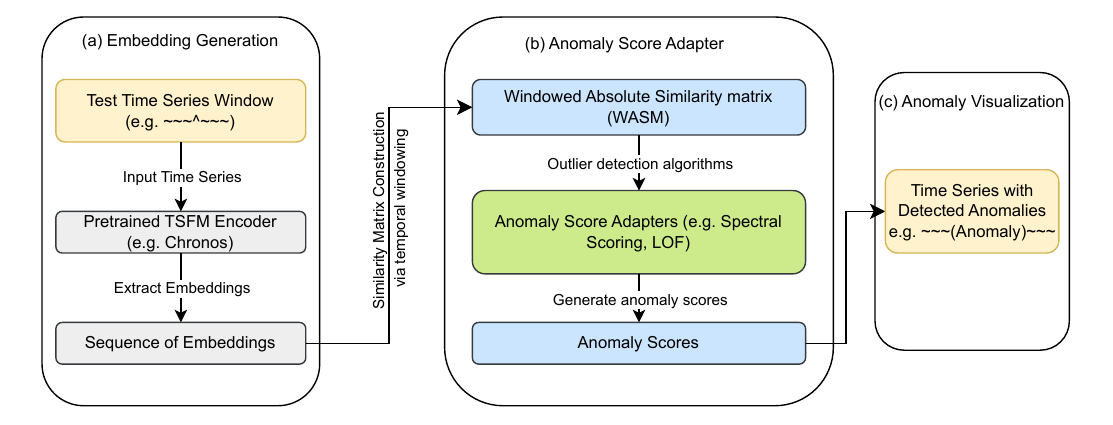} 

\caption{Architecture of the proposed anomaly detection system using TSFM embeddings and an outlier detection algorithm.}
\label{fig:architecture}
\end{figure*}

Building on the above mentioned motivations, we hereby introduce \textbf{THEMIS}: Unlocking Pre\textit{\textbf{T}}rained Knowledge wit\textbf{h} Foundation Model \textit{\textbf{Em}}beddings for Anomaly Detection in T\textit{\textbf{i}}me \textit{\textbf{S}}eries, a novel anomaly detection framework for time series, which strategically uses the pretrained knowledge encapsulated within TSFMs, enabling us to utilise the powerful representation learning capabilities of TSFMs, while decoupling the task of anomaly scoring. Instead of relying on naive direct anomaly predictions or simple thresholded forecast errors, THEMIS leverages the generic embeddings these models generate during large-scale pretraining and applies specialized outlier detection algorithms on those representations. A key practical strength of THEMIS is its robustness to hyperparameters, significantly reducing the need for exhaustive tuning and enhancing its reliability, especially where labeled anomalies are scarce.
As shown in Figure \ref{fig:architecture}, THEMIS extracts meaningful embeddings first from the encoder of the Chronos foundation model for a given time series window. It then applies a robust outlier detection algorithm to those embeddings so as to detect anomalies. Our investigations consider local outliers through unsupervised algorithms such as Local Outlier Factor (LOF)\cite{Breunig2000} and Spectral Decomposition on the self-similarity matrix, which have proved to be very effective. More precisely, our experiments show that Chronos embeddings combined with Spectral Decomposition-based outlier detection result in \textbf{state-of-the-art (SOTA)} results on the well-known MSL (Mars Science Laboratory) dataset and competitive performance on a challenging benchmark like the SMAP (Soil Moisture Active Passive) dataset. This method provides better detection accuracy, enhanced interpretability, and as previously highlighted, robustness to hyperparameter variations.

The main contributions of our work are as follows:
\begin{itemize}
\item We introduce \textbf{THEMIS}, a novel and modular framework for time series anomaly detection that successfully combines pretrained embeddings from the Chronos foundation model with established outlier detection techniques, including Local Outlier Factor and Spectral Decomposition on the self-similarity matrix. This framework is modular, and has potential to detect subtler anomalies with the evolution of TSFMs.
\item We show that THEMIS achieves SOTA performance, especially when pairing Chronos embeddings with Spectral Decomposition, on the MSL benchmark dataset, while also performing strongly on SMAP, even outperforming models explicitly trained for anomaly detection.
\item The framework, due to the rich representations learned by the foundation model and the kind of outlier detection methods applied, inherently possesses benefits such as robustness to hyperparameter variations and interpretability.
\end{itemize}

The remainder of this paper is organised as follows. Section 2 provides a review of the related work in time series forecasting-based anomaly detection, time series foundation models, how embeddings are used to perform anomaly detection, and relevant outlier detection techniques. Section 3 details our proposed methodology, THEMIS, starting with the overall framework, followed by the Anomaly Score Adapter which includes discussions on Spectral Residual Scoring, Local Outlier Factor (LOF), Score Normalization, Mean Similarity Scoring, Trimmed Top-k Similarity. Section 4 presents our experimental evaluation, including the setup, benchmark datasets, comprehensive results, and key observations. Finally, Section 5 concludes the paper with a summary of our findings and outlines promising directions for future research.

\section{Related Work}


\subsection{Forecasting-Based Anomaly Detection}

A prominent class of time series anomaly detection methods leverages forecasting models to define normal behavior. Classical techniques such as ARIMA, Exponential Smoothing, and Prophet identify anomalies by quantifying deviations between predicted and observed values~\cite{Taylor2018}. This approach assumes that well-predicted patterns are normal, and large residuals signify anomalies. Recent work has explored deep learning-based forecasters, including LSTMs~\cite{Malhotra2015} and Transformer-based models~\cite{Wen2022transformers}, which capture higher-order temporal dependencies. Despite their expressive capacity, these methods fundamentally rely on residual-based anomaly scoring, akin to classical approaches. A persistent challenge across all forecasting-based methods is the need to define error thresholds that separate anomalies from noise~\cite{Chalapathy2019survey}, often requiring manual tuning or domain-specific heuristics, which hinders generalization.


\subsection{Time Series Foundation Models}

Foundation models, characterized by large-scale pretraining on diverse datasets for broad transferability~\cite{Bommasani2021foundation}, have recently gained momentum in the time series domain. Notable examples include TimeGPT~\cite{Garza2023}, TEMPO~\cite{cao2024tempo} and Chronos~\cite{Ansari2024}, which demonstrate strong zero-shot and few-shot forecasting performance across heterogeneous time series benchmarks. Chronos adapts the T5 architecture by discretizing continuous time series into token sequences and pretraining on a large corpus of public data. In this work, \textsc{Themis} directly utilizes frozen Chronos encoder embeddings for anomaly detection, leveraging the inductive biases and generalizable structure learned during pretraining to distinguish between normal and anomalous temporal patterns.


\subsection{Anomaly Detection with Embeddings}
Embedding-based approaches have proven highly effective for anomaly detection across domains. Autoencoders, for instance, are trained to reconstruct normal instances, with high reconstruction errors indicating potential anomalies~\cite{An2015}. Beyond reconstruction, representation learning techniques inspired by NLP—such as Word2Vec—have been adapted to model categorical event sequences, where infrequent or unusual sequences are treated as anomalies~\cite{Choi2018mime}. More recently, contrastive self-supervised methods have gained prominence by learning discriminative embeddings that cluster normal patterns while isolating anomalies in sparse regions of the representation space~\cite{Shen2020,Tack2020csi}.

\textsc{Themis} aligns with this embedding-centric paradigm but diverges by leveraging general-purpose time series embeddings from Chronos, a foundation model pretrained across diverse domains. This removes the need for task-specific representation learning, enabling zero-shot anomaly detection without dataset-specific reconstruction or contrastive training.

\subsection{Outlier Detection Techniques for Anomaly Detection}

The anomaly detection component in THEMIS uses well-known outlier detection techniques on similarity matrices created from embeddings of foundation models. The two main techniques explored are Spectral Decomposition and the Local Outlier Factor.

Spectral Decomposition attempts to uncover latent data structures through eigendecomposition of a similarity matrix $S \in \mathbb{R}^{n \times n}$ into  \[  S = Q \Lambda Q^\top \], where $Q$ contains the eigenvectors and $\Lambda$ is a diagonal matrix of eigenvalues. The data is then projected into a lower-dimensional subspace defined by the top-k eigenvectors, thus capturing dominant patterns. We define anomalies to be data points that do not fit into this principal subspace very well, often manifesting either through low projection norms or high reconstruction errors. This approach assumes that normal data forms coherent structures in the spectral space, which anomalies disrupt. These methods have been widely applied in clustering \cite{ng2002spectral} and graph-based anomaly detection \cite{akoglu2015graph}. In THEMIS, spectral analysis will be performed upon similarity matrices derived from the foundation model embeddings.

The Local Outlier Factor algorithm \cite{Breunig2000}, on the other hand, evaluates the degree of isolation of a point by comparing its local density with the local densities of its neighbors. Points in locations whose density are much less than that of their vicinity are labeled with a high LOF score, and thus are treated as potential anomalies; points in areas with uniform density, on the other hand, mostly have LOF scores around one. LOF lends support in detection problems over complicated, high-dimensional embedding spaces where global density-based assumptions are unreliable.

\section{Methodology}

Given a univariate time series $\mathcal{D}_{\text{test}} = (x_1, x_2, \dots, x_T) \in \mathbb{R}^{T}$, where $T$ denotes the number of time steps, the objective of anomaly detection is to produce a sequence of binary labels $\hat{\mathbf{y}} = (y_1, y_2, \dots, y_T)$, where $y_t \in \{0,1\}$ indicates whether the observation $x_t$ at time $t$ is anomalous.

\subsection{THEMIS Framework}
We propose \textbf{THEMIS}, a novel zero-shot anomaly detection framework that leverages the representational capacity of forecasting foundation models to address the label inefficiency in AD. THEMIS is designed to operate in a plug-and-play fashion, enabling anomaly detection on unseen time series datasets without requiring any task-specific training or fine-tuning. This design ensures high adaptability and makes THEMIS a scalable and domain-agnostic solution.

THEMIS assumes access to a pretrained foundation model $\mathcal{F}$ trained on a diverse collection of univariate and multivariate time series datasets $\mathcal{D} = \{\mathcal{D}^{(i)}\}_{i=1}^{M}$, where each $\mathcal{D}^{(i)} = (x^{(i)}_1, \dots, x^{(i)}_{T^{(i)}}) \in \mathbb{R}^{T^{(i)}}$ consists of $T^{(i)}$ sequential observations. These datasets may originate from heterogeneous domains and exhibit varying temporal dynamics. THEMIS utilizes $\mathcal{F}$ as a frozen encoder, thereby decoupling representation learning from anomaly scoring, and benefiting from the generalizable features learned during forecasting pretraining.

At inference time, THEMIS uses a frozen forecasting foundation model—specifically, Chronos~\cite{Ansari2024} in our experiments—to encode fixed-length sliding windows from the input series into a high-dimensional embedding space. These embeddings capture rich temporal structure and contextual dependencies across time, learned via next-step forecasting pretraining.

\subsection{Temporal Windowing and Similarity Matrix Construction}

To balance computational efficiency with temporal expressiveness, THEMIS adopts a structured windowing approach to process long test sequences. Given a test time series $\mathcal{D}_{\text{test}} = (x_1, x_2, \dots, x_T) \in \mathbb{R}^{T}$, we follow the architectural recommendations of the Chronos foundation model~\cite{Ansari2024} and segment the input using fixed-length sliding windows of size $L = 512$, which corresponds to the context length used during Chronos pretraining.

Each such window $(x_{t}, x_{t+1}, \dots, x_{t+L-1})$ is passed to the frozen Chronos encoder $\mathcal{F}$, which outputs a contextual embedding tensor $\mathbf{Z}_t \in \mathbb{R}^{L \times d}$, where $d = 768$ is the embedding dimension. This results in a sequence of $L$ dense representations per window, with each embedding encoding local temporal patterns and dependencies captured during pretraining.

To construct a similarity representation without incurring quadratic memory costs over the entire series, we divide the full sequence of embeddings into non-overlapping \textit{embedding batches}, each containing $B$ windows. For each batch, we aggregate the embeddings into a matrix $\mathbf{Z}_{\text{batch}} \in \mathbb{R}^{B \cdot L \times d}$ and compute a self-similarity matrix $\mathbf{S} \in \mathbb{R}^{B \cdot L \times B \cdot L}$, where each entry $\mathbf{S}[i, j]$ quantifies the similarity between embeddings $\mathbf{z}_i$ and $\mathbf{z}_j$ from the current batch.

We compute similarity using cosine similarity followed by an element wise absolute-value operation to retain both positively and negatively correlated temporal patterns:
\[
\mathbf{S}[i,j] = \left| \frac{\langle \mathbf{z}_i, \mathbf{z}_j \rangle}{\|\mathbf{z}_i\|_2 \cdot \|\mathbf{z}_j\|_2} \right|.
\]
We denote the resulting matrix as the \textit{Windowed Absolute Similarity Matrix} (WASM), and refer to it henceforth as \(\mathbf{S}\). WASM encodes localized structural relationships across multiple contextual spans by computing the element-wise absolute similarity among embeddings extracted from sliding context windows. This matrix \(\mathbf{S}\) serves as the unified input to the subsequent anomaly score adapters described in the following sections. This design ensures that \textsc{Themis} remains both computationally scalable and contextually expressive, enabling effective detection of diverse anomalous patterns in long time series.

\subsection{Anomaly Score Adapters}
\label{sec_anomaly_score_adapters}

The core of \textbf{THEMIS} is its modular anomaly scoring framework, leveraging diverse model-agnostic techniques operating atop frozen foundation model embeddings. Given the Windowed Absolute Similarity Matrix (WASM), denoted as $\mathbf{S} \in \mathbb{R}^{B \cdot L \times B \cdot L}$, our objective is to derive anomaly scores $s_t$ for each timestamp $t \in \{1, \dots, T\}$, corresponding to each data point in the test time series $\mathcal{D}_{\text{test}}$.

We investigate several complementary scoring strategies:

\paragraph{(i) Spectral Residual Scoring}  
We perform eigendecomposition of the WASM,
\[
\mathbf{S} = \mathbf{Q}\boldsymbol{\Lambda}\mathbf{Q}^\top,
\]
where \(\boldsymbol{\Lambda} = \text{diag}(\lambda_1,\dots,\lambda_{B\cdot L})\) (ascending) and \(\mathbf{Q} = [\mathbf{q}_1,\dots,\mathbf{q}_{B\cdot L}]\). Retaining the top-\(k\) eigenvectors yields 
\[
\mathbf{E} = [\mathbf{q}_{B\cdot L - k + 1}, \dots, \mathbf{q}_{B\cdot L}].
\]
The anomaly score for each point \(t\) is 
\[
s_t = 1 - \frac{\|\mathbf{e}_t\|_2}{\max_j \|\mathbf{e}_j\|_2},
\]
so that points poorly aligned with the dominant subspace receive higher scores~\cite{akoglu2015graph,ng2002spectral}.

\paragraph{(ii) Local Outlier Factor (LOF) Scoring}  
We convert \(\mathbf{S}\) to a distance matrix \(\mathbf{D}\) via \(D_{ij} = \max(\mathbf{S}) - S_{ij}\). For each point \(t\), the local reachability density is
\[
\mathrm{LRD}_k(t) = \Bigl(\tfrac{1}{|N_k(t)|} \sum_{j \in N_k(t)} \max\{D_{tj},\,\mathrm{k\text{-}dist}(j)\}\Bigr)^{-1},
\]
and the LOF score is
\[
\mathrm{LOF}_k(t) = \tfrac{1}{|N_k(t)|}\sum_{j \in N_k(t)} \frac{\mathrm{LRD}_k(j)}{\mathrm{LRD}_k(t)},
\]
with higher values indicating points in sparser neighborhoods~\cite{Breunig2000}.

\paragraph{(iii) Mean Similarity Scoring}  
For each point \(t\), compute the average similarity 
\[
\mu_t = \tfrac{1}{B\cdot L - 1} \sum_{j \neq t} S_{tj},
\]
and define 
\[
s_t = 1 - \mu_t.
\]
This simple measure treats points with low global alignment as anomalous.

\paragraph{(iv) Trimmed Top-\(k\) Similarity Mean}  
To mitigate the influence of extreme similarity values, we compute a trimmed top-\(k\) mean of each point’s similarity scores by discarding a small fraction of smallest and largest similarities before averaging the remaining top-\(k\) values. The resulting score 
\[
s_t = 1 - t_t
\]
highlights points that are poorly aligned with their most relevant neighbors. Despite its simplicity, this method provides a strong baseline for comparing against more complex adapter strategies. Full derivations and parameter settings are provided in Appendix~\ref{appendix:adapters}.








\paragraph{Score Normalization}
Anomaly scores from each scoring method are standardized via min-max normalization:

$$
s_t = \frac{s_t - \min_j s_j}{\max_j s_j - \min_j s_j + \varepsilon},
$$

with a small constant $\varepsilon$ (e.g., $10^{-9}$) ensuring numerical stability. This common normalization facilitates comparative interpretation across all scoring methods.




\subsection{Anomaly Criterion}
\label{sec:anomaly-criterion}

During inference, \textsc{THEMIS} computes unsupervised anomaly scores per timestep via spectral embedding norms derived from the foundation model-induced similarity structure, effectively capturing deviations from the dominant manifold. Alternative scoring strategies, including Local Outlier Factor (LOF) and top-$k$ trimmed similarity means, are also supported, though spectral scoring consistently demonstrates superior empirical performance. Once anomaly scores are computed, we adopt the SPOT algorithm~\cite{spot} to derive an adaptive decision threshold. SPOT models the tail distribution of scores using a Generalized Pareto Distribution and provides a statistically principled threshold $\delta$. A time point is flagged as anomalous if its score exceeds $\delta$. This thresholding strategy has been used in prior state-of-the-art methods~\cite{dada,wang2023drift,su2019robust}, and complements the zero-shot, unsupervised nature of \textsc{Themis}.


\section{Experiments}
\begin{table*}[ht]
\centering
\begin{tabular}{|c|ccc|ccc|ccc|}
\hline
\textbf{Dataset} & \multicolumn{3}{c|}{\textbf{MSL}} & \multicolumn{3}{c|}{\textbf{SMAP}} & \multicolumn{3}{c|}{{\textbf{SWaT}$^*$}} \\

\hline
Metric & Precision & Recall & F1 & Precision & Recall & F1 & Precision & Recall & F1 \\
\hline
OCSVM       & 50.26 & 99.86 & 66.87 & 41.05 & 69.37 & 51.58 & 56.80 & 98.72 & 72.11 \\
PCA         & 52.69 & 98.33 & 68.61 & 50.62 & 98.48 & 66.87 & 62.32 & 82.96 & 71.18 \\
HBOS        & 59.25 & 83.32 & 69.25 & 41.54 & 66.17 & 51.04 & 54.49 & 91.35 & 68.26 \\
LOF         & 49.89 & 72.18 & 59.00 & 47.92 & 82.86 & 60.72 & 53.20 & 96.73 & 68.65 \\
IForest     & 53.87 & 94.58 & 68.65 & 41.12 & 68.91 & 51.51 & 53.03 & 99.95 & 69.30 \\
LODA        & 57.79 & 95.65 & 72.05 & 51.51 & 100.00 & 68.00 & 56.30 & 70.34 & 62.54 \\
AE          & 55.75 & 96.66 & 70.72 & 39.42 & 70.31 & 50.52 & 54.92 & 98.20 & 70.45 \\
DAGMM       & 54.07 & 92.11 & 68.14 & 50.75 & 96.38 & 66.49 & 59.42 & 92.36 & 72.32 \\
LSTM        & 58.82 & 14.68 & 23.49 & 55.25 & 27.70 & 36.90 & 49.99 & 82.11 & 62.15 \\
BeatGAN     & 55.74 & 98.94 & 71.30 & 54.04 & 98.30 & 69.74 & 61.89 & 83.46 & 71.08 \\
Omni        & 51.23 & 99.40 & 67.61 & 52.74 & 98.51 & 68.70 & 62.76 & 82.82 & 71.41 \\
CAE-Ensemble & 54.99 & 93.93 & 69.37 & 62.32 & 64.72 & 63.50 & 62.10 & 82.90 & 71.01 \\
MEMTO       & 52.73 & 97.34 & 68.40 & 50.12 & 99.10 & 66.57 & 56.47 & 98.02 & 71.66 \\
A.T.        & 51.04 & 95.36 & 66.49 & 56.91 & 96.69 & 71.65 & 53.63 & 98.27 & 69.39 \\
DCdetector  & 55.94 & 95.53 & 70.56 & 53.12 & 98.37 & 68.99 & 53.25 & 98.12 & 69.03 \\
SensitiveHUE& 55.92 & 98.95 & 71.46 & 53.63 & 98.37 & 69.42 & 58.91 & 91.71 & 71.74 \\
D3R         & 66.85 & 90.83 & 77.02 & 61.76 & 92.55 & 74.09 & 60.14 & 97.57 & 74.41 \\
ModernTCN   & 65.94 & 93.00 & 77.17 & 69.50 & 65.45 & 67.41 & 59.14 & 89.22 & 71.13 \\
GPT4TS      & 64.86 & 95.43 & 77.23 & 63.52 & 90.56 & 74.67 & 56.84 & 91.46 & 70.11 \\
DADA & 68.70 & 91.51 & 78.48 & 65.85 & 88.25 & \textbf{75.42} & 61.59 & 94.59 & \textbf{74.60} \\
\hline
\textbf{THEMIS(Ours)} & 70.51 & 89.25 & \textbf{78.78} & 61.14 & 91.21 & 73.21 & 56.00 & 99.20 & 71.59 \\
\hline
\end{tabular}
\caption{Precision, Recall, and F1 scores of various models across MSL, SMAP, and SWaT$^*$ datasets. All results are in \%.}
\label{tab:main_benchmark}
\end{table*}

\begin{table*}[ht]
\centering
\begin{tabular}{|c|ccc|ccc|ccc|}
\hline
\textbf{Dataset} & \multicolumn{3}{c|}{\textbf{MSL}} & \multicolumn{3}{c|}{\textbf{SMAP}} & \multicolumn{3}{c|}{\textbf{SWaT}$^*$} \\

\hline
Metric & Precision & Recall & F1 & Precision & Recall & F1 & Precision & Recall & F1 \\
\hline
mean       & 62.31 & 99.27 & 76.56 & 58.32 & 88.15 & 70.20 & 53.95 & 99.98 & 70.09 \\
trimmed-mean        & 63.68 & 97.92 & 77.18 & 54.08 & 98.70 & 69.87 & 53.60 & 99.99 & 69.79 \\
LOF         & 58.64 & 93.91 & 72.20 & 60.70 & 93.21 & 73.52 & 51.10 & 71.02 & 59.44 \\
\hline
\textbf{THEMIS} & 70.51 & 89.25 & \textbf{78.78} & 61.14 & 91.21 & 73.21 & 56.00 & 99.20 & 71.59 \\
\hline
\end{tabular}
\caption{Precision, Recall, and F1 scores of Anomaly Score Adapter on MSL, SMAP, SWaT$^*$. All results are in \%.}
\label{tab:themis_baselines}
\end{table*}

\subsection{Experimental Settings}
\label{sec:experimental-settings}
\paragraph{Datasets.}
To ensure robust generalization and validate the zero-shot capabilities of our proposed framework, we utilize the \texttt{Chronos} foundation model \cite{Ansari2024}, pretrained exclusively on a diverse corpus comprising approximately 55 univariate time series datasets spanning multiple real-world domains, including finance, healthcare, energy, manufacturing, and sensor networks. Notably, none of the evaluation benchmarks employed in our experiments are present within the pretraining corpus. Given \texttt{Chronos}' univariate forecasting constraint, we evaluate our framework on single-channel subsets of three established multivariate anomaly detection benchmarks: MSL, SMAP\cite{smap}, and SWaT\cite{swat}. Specifically, for MSL and SMAP, we follow the channel-selection methodology in\cite{dada}, while for SWaT, we select the first available channel, denoted henceforth as SWaT$^*$. These datasets respectively represent spacecraft telemetry, mechanical systems, and cyber-physical infrastructures, ensuring comprehensive coverage across diverse temporal anomaly scenarios and are well-established benchmarks in the community. Additional dataset and implementation details are provided in Appendix~\ref{appendix:datasets} and Appendix \ref{appendix:implementation} respectively.
\paragraph{Baselines.}
We compare \textsc{Themis} with a comprehensive set of strong baselines, consistent with prior literature~\cite{dada,anomalytransformer,dcdetector}. These include classical methods such as OCSVM~\cite{ocsvm}, PCA~\cite{pca}, LOF~\cite{Breunig2000}, and IForest~\cite{iforest}, as well as modern neural and transformer-based approaches such as AutoEncoder~\cite{autoencoders_sakurada}, DAGMM~\cite{dagmm}, OmniAnomaly~\cite{omnianomaly}, BeatGAN~\cite{beatgan}, Anomaly Transformer~\cite{anomalytransformer}, MEMTO~\cite{memto}, DCdetector~\cite{dcdetector}, D3R~\cite{d3r}, GPT4TS~\cite{gpt4ts} and recent DADA ~\cite{dada}, which proposes a diffusion-based anomaly detection mechanism. All baselines are under standardized conditions, ensuring fairness through consistent metrics and thresholding protocols aligned closely with those outlined in recent literature \cite{dada}.
\paragraph{Metrics.}
We follow the evaluation protocol advocated by recent studies~\cite{huet2022local,dada}, which highlight the limitations of the widely used Point Adjustment (PA) heuristic. PA often inflates performance metrics by marking an entire anomaly segment as correctly detected if any single point within it is identified, thereby leading to an over-optimistic assessment. To address this, we adopt the \emph{affiliation-based} F1 score (F1)~\cite{huet2022local,dada,dcdetector,wang2023drift}, a temporally-aware metric that measures the alignment between predicted and true anomaly segments using affiliated precision (P) and recall (R). Given the inherent sensitivity of both precision and recall to the choice of threshold, evaluating models solely based on one of these metrics provides an incomplete picture of detection quality. Therefore, in line with recent practice~\cite{dada,dcdetector,anomalytransformer}, we primarily report the F1 score as our main evaluation criterion. In all tables, the best results are highlighted in bold.

\begin{figure*}[ht]
    \begin{minipage}{0.5\textwidth}
    \includegraphics[width=\linewidth]{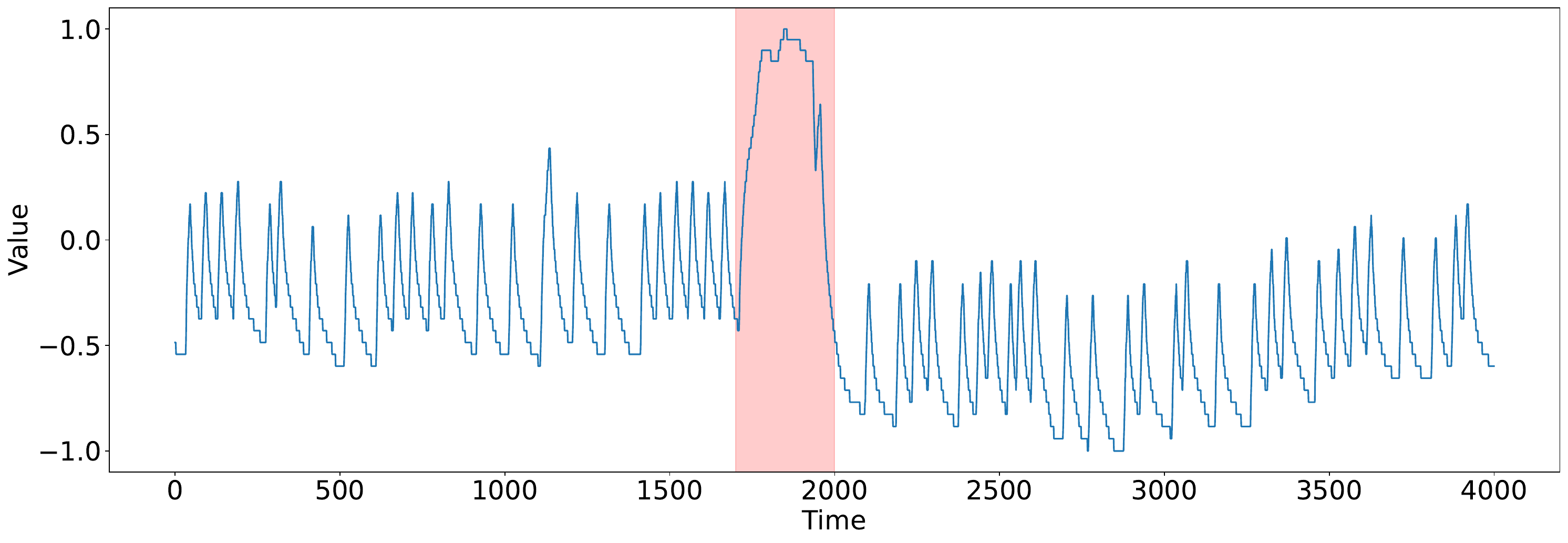}\par 
    \end{minipage}
     \begin{minipage}{0.5\textwidth}
    \includegraphics[width=\linewidth]{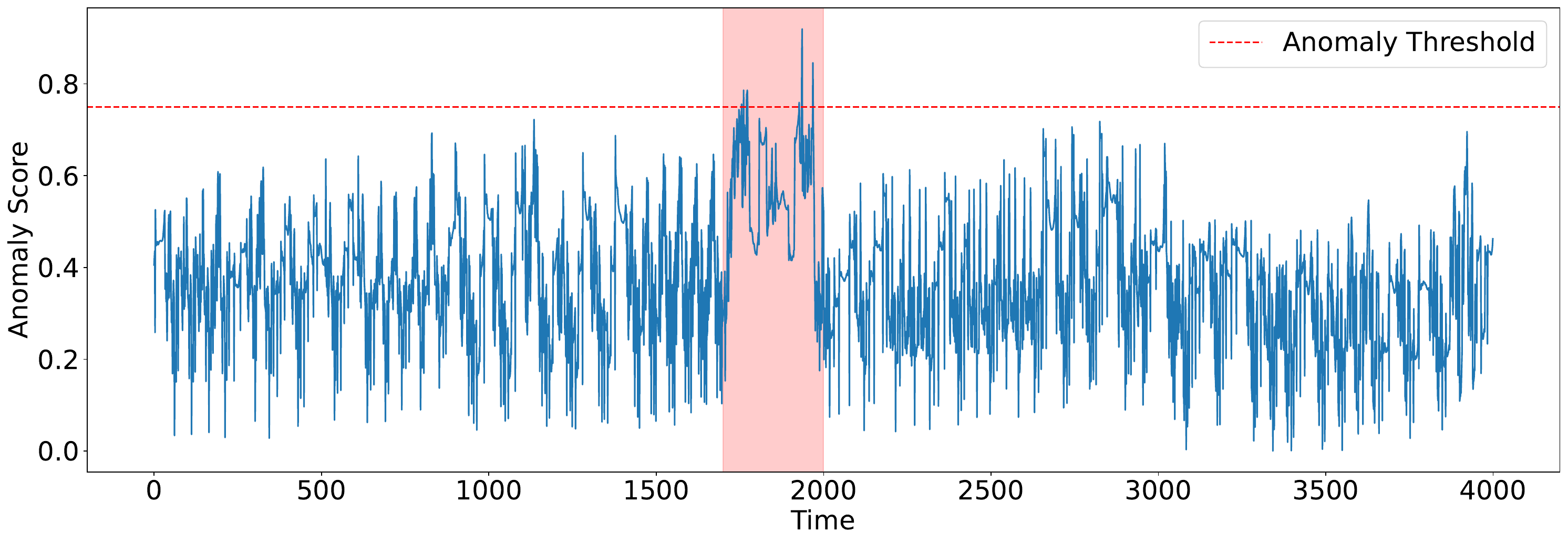}\par  
    \end{minipage}
   \caption{Visualization of anomaly scores on a segment from the SMAP dataset. \textbf{Left:} Raw time series input with ground truth anomaly regions shaded in red. \textbf{Right:} Corresponding anomaly scores generated by \textsc{Themis}. The red horizontal line denotes the threshold. Anomalous regions are accurately identified with high anomaly scores, while non-anomalous regions exhibit low scores. Anomalous regions are shaded in red.}
    \label{fig:msl_plot}
\end{figure*}
\begin{figure*}[ht]
    \begin{minipage}{0.5\textwidth}
    \centering
    \includegraphics[width=.8\linewidth]{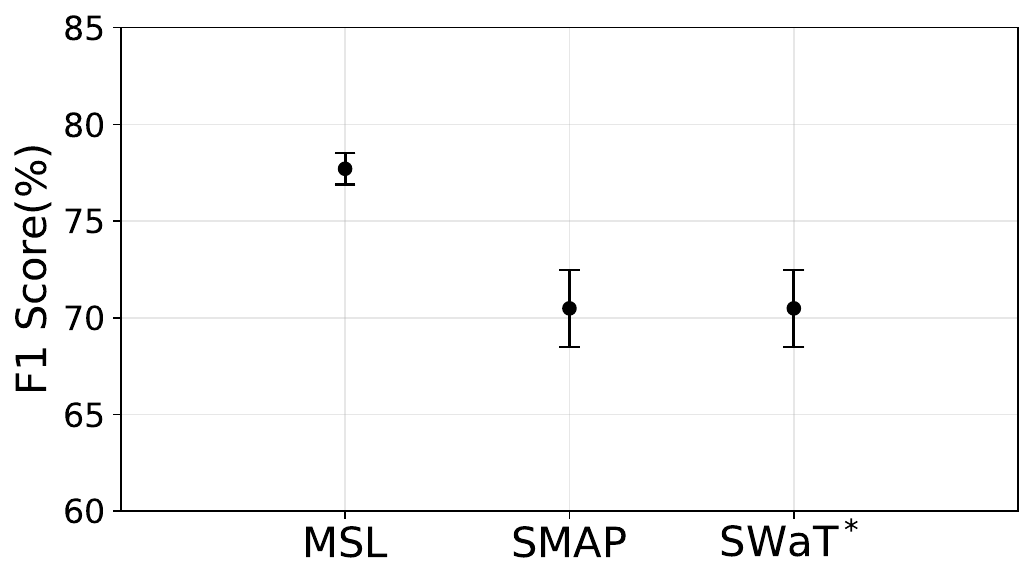}\par 
    \end{minipage}
     \begin{minipage}{0.5\textwidth}
     \centering
    \includegraphics[width=.5\linewidth, keepaspectratio]{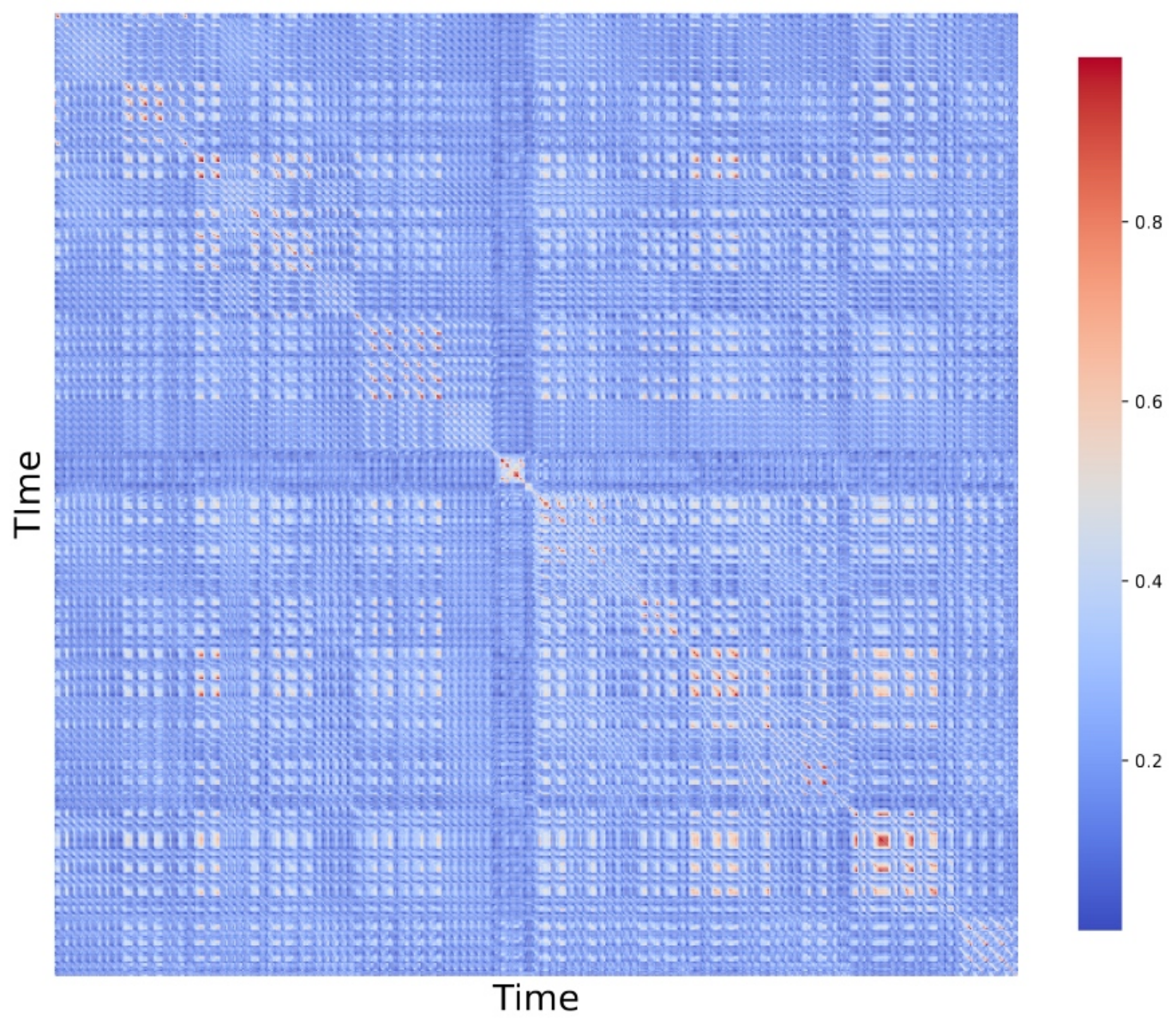}\par  
    \end{minipage}
    \caption{\textbf{Left:}Error bar plots show the mean F1-score and standard deviation for different values of the spectral scoring hyperparameter \(k\) (number of top eigenvectors used) on the MSL, SMAP, and SWaT$^*$ datasets. \textsc{Themis} consistently achieves stable and high F1-scores across a wide range of \(k\) values. \textbf{Right:} Heatmap of the similarity matrix $\mathbf{S}$ for segment of SMAP shown in Figure \ref{fig:msl_plot}(\textbf{Left}). Anomalous points exhibit low similarity to normal regions.
  }
    \label{fig:errorbarplot}
\end{figure*}

\subsection{Main Results}
We evaluate \textsc{Themis} under a rigorous zero-shot anomaly detection setting, where the Chronos foundation model is pretrained solely on large-scale forecasting datasets without exposure to any downstream anomaly detection data or labels. This setup reflects a realistic, label-sparse deployment scenario and provides a compelling testbed for assessing generalization capabilities; notably, benchmarks such as MSL, SMAP, and SWaT are not included in the pretraining corpus, ensuring strict separation between training and test domains.

Across experiments, the spectral residual scoring adapter consistently yields the highest anomaly detection performance in both point-wise and contextual tasks. In particular, on the MSL benchmark, \textbf{THEMIS}—i.e., Chronos embeddings coupled with spectral residual scoring—surpasses several supervised and unsupervised baselines without any task-specific tuning or labeled data. Henceforth, unless otherwise specified, \textbf{THEMIS} refers to this combination.


\textbf{Performance Across Benchmarks.} As summarized in Table~\ref{tab:main_benchmark}, \textsc{Themis} achieves state-of-the-art results on the MSL dataset, attaining an F1-score of \textbf{78.78\%}, surpassing all 19 baselines including advanced neural approaches such as GPT4TS (\underline{77.23\%}) and DADA (\underline{78.48\%}). On SMAP and SWaT, \textsc{Themis} maintains highly competitive performance (73.21\% and 71.59\% F1 respectively), closely approaching or exceeding methods that are explicitly trained on these benchmarks. These results are particularly notable given that \textsc{Themis} operates in a univariate, zero-shot regime, without any task-specific tuning or labeled anomaly data.

\textbf{Efficacy of Anomaly Score Adapters.} A central component of \textsc{Themis} is its modular anomaly scoring interface, which enables plug-and-play integration of multiple outlier scoring strategies applied to Chronos-derived embeddings. To assess this component's flexibility and robustness, we conduct an ablation study (Table~\ref{tab:themis_baselines}) comparing four adapters: (i) \emph{mean similarity scoring}, (ii) \emph{trimmed top-k mean}, (iii) \emph{Local Outlier Factor (LOF)}, and (iv) our proposed \emph{spectral residual scoring}.

Across datasets, we observe that spectral residual scoring consistently yields the best or near-best performance. For instance, on MSL, it achieves an F1-score of 78.78\%, significantly outperforming LOF (72.20\%) and trimmed mean (77.18\%). While simple averaging (mean similarity) yields reasonable results, the spectral method demonstrates superior capacity to capture global irregularities in the similarity structure of embeddings, especially in complex multivariate settings. This validates the utility of spectral decomposition for capturing subtle deviations in temporal dynamics that simpler methods may overlook.

\textbf{Robustness to Hyperparameters.} One of the practical strengths of \textsc{Themis} lies in its robustness to hyperparameter variations, particularly within its spectral residual scoring component. As illustrated in Figure~\ref{fig:errorbarplot} (\textbf{Left}), the F1 scores remain consistently high across a wide range of values for the primary hyperparameter—the number of eigenvectors $k$ retained from the similarity matrix. We evaluate $k \in \{2, 5, 10, 15, 20\}$ and observe minimal degradation in performance across this spectrum. This empirical stability significantly reduces the need for exhaustive hyperparameter tuning, which is especially advantageous in real-world scenarios where labeled anomalies are scarce or tuning budgets are constrained. Such resilience not only enhances the practicality of \textsc{Themis} but also underscores the reliability of its spectral scoring approach across diverse data regimes.

\textbf{Zero-Shot Generalization Without Anomaly Supervision.} Perhaps most critically, \textsc{Themis} achieves these results without any form of supervised learning for anomaly detection. The Chronos model is trained solely on forecasting objectives and yet yields rich, semantically meaningful embeddings suitable for detecting anomalies across a diverse set of domains. This decouples the need for costly anomaly annotations and illustrates that strong inductive biases in forecasting-based foundation models can be harnessed for high-quality, domain-agnostic anomaly detection.

\textbf{Visual Analysis.} To qualitatively assess the interpretability and practical efficacy of \textsc{Themis}, we present a representative example from the SMAP dataset in Figure~\ref{fig:msl_plot}. The left panel displays the raw time series, while the right panel shows the anomaly scores generated by \textsc{Themis} using the spectral residual adapter. The model accurately assigns elevated scores to anomalous intervals, exhibiting strong alignment with the ground truth despite operating in a zero-shot setting.

Further structural insight is provided in Figure~\ref{fig:errorbarplot} (\textbf{Right}), which visualizes the corresponding similarity matrix \( \mathbf{S} \) derived from the same time series segment. Here, normal regions maintain high similarity with one another, whereas anomalous points exhibit diminished similarity to the rest of the series, reflecting their structural disconnect in the learned representation space. Additional visual examples across diverse anomaly patterns from the NAB \cite{nab_dataset} artificial anomaly dataset are provided in Appendix~\ref{appendix:visual_analysis}, further illustrating the robustness and interpretability of \textsc{Themis}.

\section{Conclusion and Future Work}
In conclusion, the paper discusses significant and persistent challenges in time series anomaly detection. We propose THEMIS, a new framework that effectively harnesses pretrained knowledge from foundation models to detect anomalies on time series data. Upon extracting embeddings from the Chronos foundation model, THEMIS applies robust outlier detection methods such as Local Outlier Factor and Spectral Decomposition on the self-similarity matrix, resulting in a modular and effective approach.

The effectiveness of this methodology is underscored by our experiments. THEMIS sets a SOTA performance on the MSL dataset and achieves highly competitive results on the SMAP dataset. Furthermore, it outperforms models directly trained for anomaly detection and inherently provides hyperparameter robustness and interpretability. This study promotes the use of pretrained representations from foundation models as a very promising general solution for time series anomaly detection. The success of THEMIS opens a potentially fruitful research avenues to investigate other foundation models and sophisticated outlier detection methods to advance anomaly detection systems' capabilities in complex real-world environments.

THEMIS’s results indicate potential for further research in multiple avenues, not limited to:
\begin{itemize}
    \item Extracting embeddings from other prominent Time Series Foundation Models (TSFMs), such as Toto \cite{cohen2025time} which may contain representations better suited for separating anomalous data.
    \item Adapt THEMIS for multivariate time series, which presents its own challenges and is quite common in real-world scenarios.
    \item Studying a wider variety of sophisticated outlier detection algorithms that might increase the framework's sensitivity and robustness.
    \item Employing higher computational resources, which enables us to consider a much larger batch size in embedding generation, potentially resulting in more thorough representations and better outlier detection, thereby making rigorous comparisons feasible and leading to detecting subtler anomalies. 
\end{itemize}


\bibliographystyle{named}
\bibliography{paper} 
\clearpage

\appendix
\section{Datasets}
\label{appendix:datasets}

\begin{table*}[ht]
\centering
\label{tab:datasets}

\begin{tabular}{lccccc}
\toprule
\textbf{Dataset} & \textbf{Domain} & \textbf{Dimension}  & \textbf{Test (labeled)} & \textbf{AR (\%)} \\
\midrule
MSL & Spacecraft & 1  & 73,729 & 10.5 \\
SMAP & Spacecraft & 1  & 427,617 & 12.8 \\
SWaT* & Water Treatment & 1 & 449,919 & 12.1 \\
\bottomrule
\end{tabular}
\caption{Summary of datasets used for evaluation. All settings are univariate, using only the first channel. AR denotes the anomaly ratio in the labeled test set.}
\end{table*}

We evaluate \textbf{THEMIS} on three widely-used time series anomaly detection benchmarks spanning spacecraft telemetry and industrial control systems. All evaluations are conducted in the univariate setting, using only the first channel from each dataset to align with the univariate design of the \texttt{Chronos} foundation model. In contrast, baseline results for SWaT are based on the full multivariate input; we denote our univariate variant as \textbf{SWaT*} to highlight this distinction.

\begin{itemize}
    \item \textbf{MSL (Mars Science Laboratory)}~\cite{smap} is a spacecraft telemetry dataset collected by NASA, containing sensor and actuator readings from the Curiosity rover. We retain only the first continuous channel following the protocol in~\cite{dada}.
    
    \item \textbf{SMAP (Soil Moisture Active Passive)}~\cite{smap} is another NASA spacecraft dataset, consisting of telemetry data used to monitor soil moisture via satellite systems. We use only the first univariate channel consistent with prior work~\cite{dada}.
    
    \item \textbf{SWaT* (Secure Water Treatment)}~\cite{swat} contains sensor data from a fully operational water treatment plant testbed. Since the original dataset is multivariate and baselines operate on the full input, we evaluate only on the first channel and denote this reduced setting as SWaT*.
\end{itemize}

\section{Implementation Details}
\label{appendix:implementation}

We provide here the key implementation details necessary for reproducing \textbf{THEMIS} under the zero-shot anomaly detection setting.

\paragraph{Foundation Model Configuration.}  
All experiments use the pretrained \texttt{Chronos} (chronos-t5-base) foundation model~\cite{Ansari2024}, trained on a corpus of approximately 55 univariate time series datasets spanning a wide array of domains, including finance, healthcare, energy, manufacturing, and environmental monitoring. The model was trained with a default context length of $L = 512$ and produces embeddings of dimension $d = 768$. We adhere strictly to these pretrained specifications during inference, using the encoder in frozen mode throughout.

\paragraph{Batch-wise Similarity Construction.}  
To construct the Windowed Absolute Similarity Matrix (WASM) $\mathbf{S} \in \mathbb{R}^{B \cdot L \times B \cdot L}$, we vary the number of sliding windows per batch, denoted by $B$, to study the trade-off between locality and global context. Specifically, we experiment with $B \in \{1, 4, 16\}$, and report all main results for $B = 16$ unless otherwise specified. This setting offers a good balance between memory efficiency and temporal expressiveness.
\paragraph{Spectral Residual Scoring.}  
For the spectral anomaly score adapter, we compute the eigendecomposition of the similarity matrix and vary the number of top eigenvectors $k \in \{2, 5, 10, 15, 20\}$ retained to construct the spectral subspace. The anomaly score is then computed from the spectral norm of each data point in this $k$-dimensional space. We observe stable performance across this range, with detailed variance visualizations presented in Figure~\ref{fig:errorbarplot}.

\paragraph{Local Outlier Factor (LOF).}  
For LOF-based scoring, we transform $\mathbf{S}$ into a distance matrix and compute local reachability densities. We explore $k$-nearest neighbor values $k \in \{5, 10, 15, 20\}$ to assess the influence of neighborhood size on local density estimation.

\paragraph{Normalization and Post-processing.}  
All scoring outputs are subjected to min-max normalization to the $[0, 1]$ range. Thresholding is performed using the SPOT algorithm~\cite{spot} on the validation split, ensuring an adaptive and distribution-aware decision boundary for test-time anomaly detection.

All reported results for each anomaly score adapter correspond to their best-performing configurations, as summarized in Table~\ref{tab:themis_baselines}. Baseline results are reported under standardized conditions, adhering to the metrics and thresholding protocols detailed in \cite{dada}, and are presented here for direct comparison. Comprehensive results across varying hyperparameter settings are provided in Appendix~\ref{appendix:experimental_results}.

\section{Experimental Results}
\label{appendix:experimental_results}
\begin{table}[H]
\label{tab:spectral_k_sweep}
\begin{tabular}{c|c|c|c}
\toprule
\textbf{\(k\)} & \textbf{MSL} & \textbf{SMAP} & \textbf{SWaT*} \\
\midrule
2  & 78.3 & 72.4 & 67.4 \\
5  & 77.3 & 73.2 & 71.6 \\
10 & 77.8 & 70.1 & 67.8 \\
15 & \textbf{78.8} & 68.2 & 69.4 \\
20 & 76.4 & 68.6 & 68.9 \\
\bottomrule
\end{tabular}
\caption{F1-scores (\%) for spectral residual scoring across different values of \(k\) eigenvectors on MSL, SMAP, and SWaT* datasets.}
\end{table}

\begin{figure}[h]
    \centering
    \includegraphics[width=0.8\linewidth]{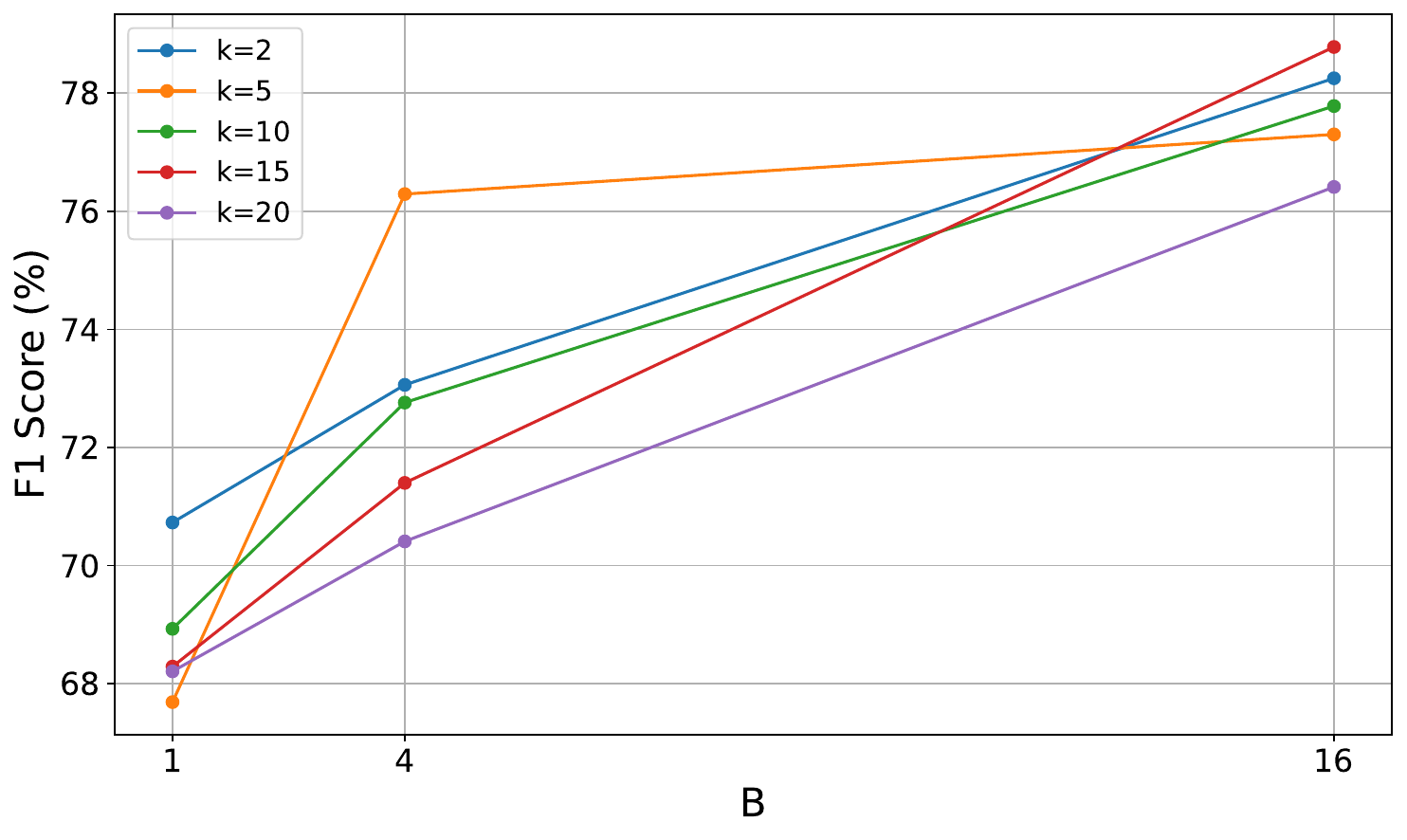}
    \caption{\textbf{F1-scores (\%) of spectral residual anomaly scoring on the MSL dataset across varying batch sizes} \(B \in \{1, 4, 16\}\) \textbf{and number of retained eigenvectors} \(k \in \{2, 5, 10, 15, 20\}\). Each curve corresponds to a fixed \(k\). Larger batch sizes consistently yield improved detection performance, indicating the benefit of broader contextual similarity modeling.}
    \label{fig:spectral_bs_k_msl}
\end{figure}
\paragraph{Spectral Sensitivity to B} Figure~\ref{fig:spectral_bs_k_msl} illustrates the impact of varying the number of sliding windows per batch \(B \in \{1, 4, 16\}\) on spectral residual anomaly detection performance across different eigenvector ranks \(k\). Across all configurations, increasing \(B\) yields consistently higher F1-scores, highlighting the benefit of aggregating broader temporal context within the similarity matrix. This trend underscores the role of batch size in enhancing structural resolution, thereby enabling more discriminative spectral embeddings for anomaly scoring.

\section{Additional Visual Analysis}
\subsection{Visual Analysis on NAB Dataset}
\label{appendix:visual_analysis}

Figures below in \ref{appendix:visual_analysis} presents qualitative examples from the NAB dataset showcasing diverse artificial anomaly patterns. Each row corresponds to a distinct time series instance exhibiting characteristic deviations.

For each example, the top-left panel displays the input time series with ground truth anomalies highlighted. The top-right panel shows the corresponding heatmap of the Windowed Absolute Similarity Matrix (WASM) $\mathbf{S}$, which encodes the pairwise similarity between temporal embeddings. Notably, anomalous regions exhibit reduced similarity to the rest of the sequence, manifesting as darker (low-value) blocks when contrasted with normal segments.

The bottom-left and bottom-right panels show anomaly scores produced by \textsc{Themis} using the spectral residual and LOF adapters, respectively. In both cases, elevated anomaly scores align well with the ground truth anomalous intervals, reflecting the model’s ability to capture both structural and local deviations.

These visualizations provide interpretability into how \textsc{Themis} exploits structural disruptions in the similarity space and reinforce the robustness of different adapter mechanisms across a variety of anomaly types.

\begin{figure*}[ht]
    \centering
    \begin{minipage}[t]{0.48\textwidth}
        \centering
        \includegraphics[width=\linewidth]{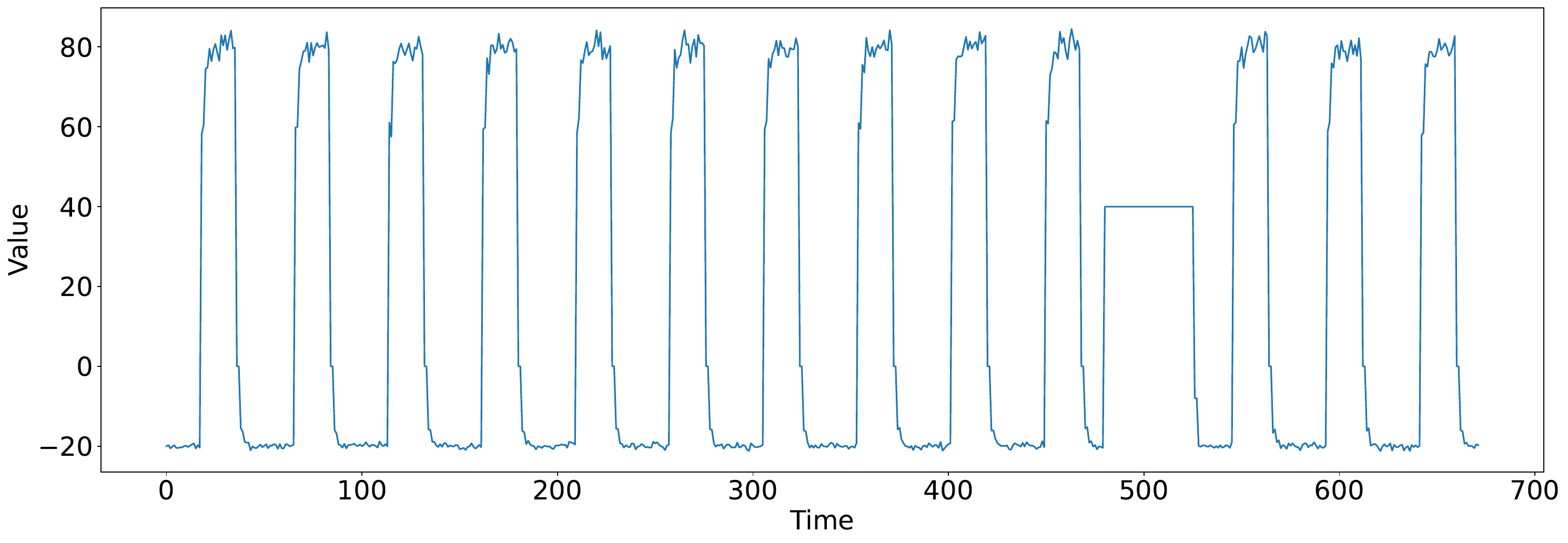}
        \subcaption{Input time series from the NAB dataset.}
    \end{minipage}%
    \hfill
    \begin{minipage}[t]{0.48\textwidth}
        \centering
        \includegraphics[width=\linewidth, height=4cm, keepaspectratio]{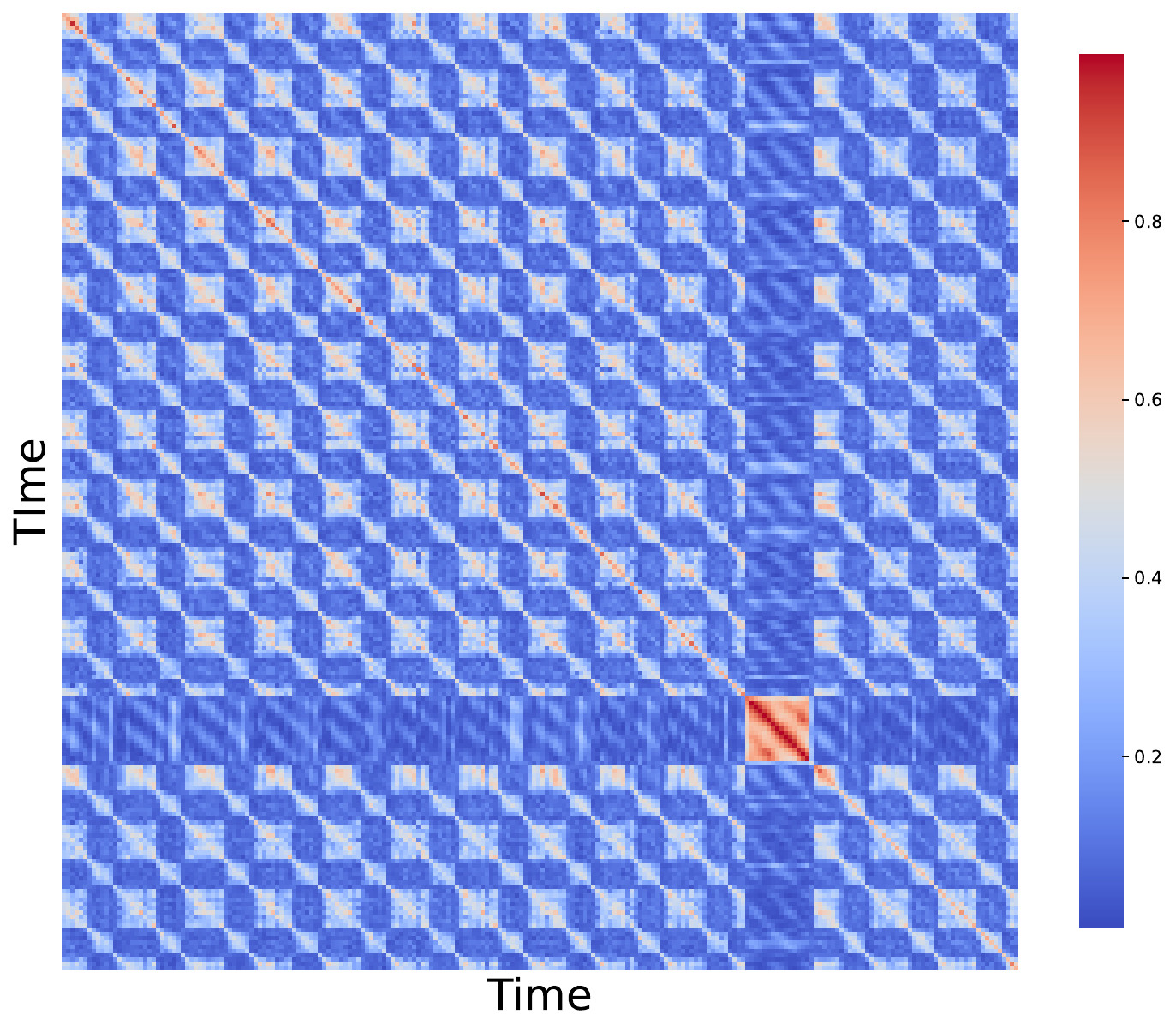}
        \subcaption{Heatmap of the similarity matrix $\mathbf{S}$. Anomalous points exhibit low similarity to normal regions.}
        \label{fig:example1_heatmap}
    \end{minipage}

    \vspace{0.5em}

    \begin{minipage}[t]{0.48\textwidth}
        \centering
        \includegraphics[width=\linewidth]{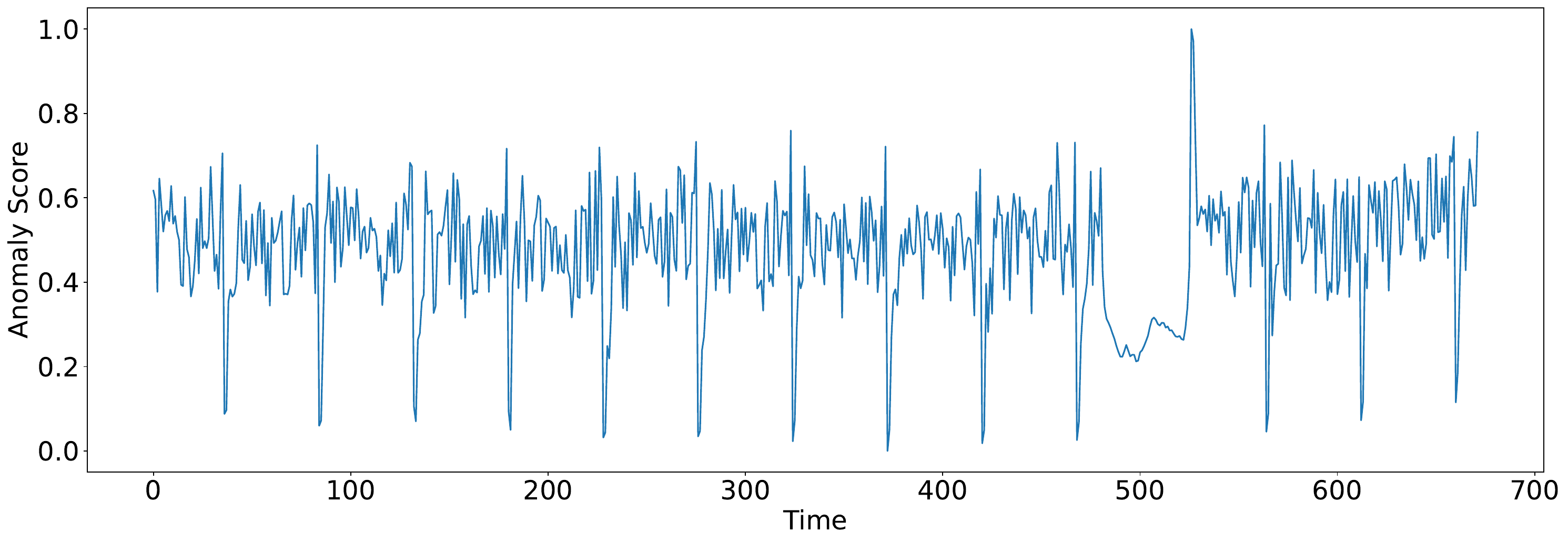}
        \subcaption{Anomaly scores from \textsc{Themis} using the spectral residual adapter.}
        \label{fig:example1_spectral}
    \end{minipage}%
    \hfill
    \begin{minipage}[t]{0.48\textwidth}
        \centering
        \includegraphics[width=\linewidth]{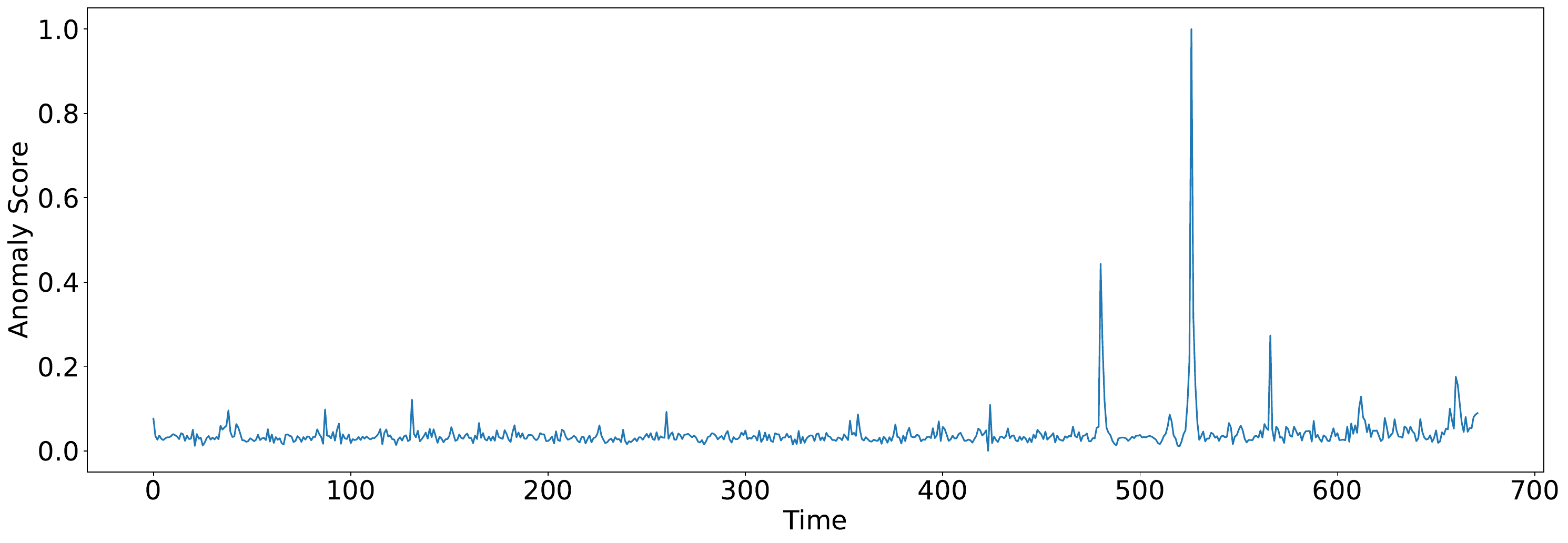}
        \subcaption{Anomaly scores from \textsc{Themis} using the LOF adapter.}
        \label{fig:example1_lof}
    \end{minipage}

    \caption{Example 3: Visual analysis on a time series from the NAB artificial anomaly dataset. \textbf{Top:} Input sequence (left) and corresponding similarity matrix (right), where anomalous points show reduced similarity to the rest of the series. \textbf{Bottom:} Anomaly scores computed by \textsc{Themis} using spectral residual (left) and LOF (right) adapters, both effectively highlighting anomalous regions.}
    \label{fig:nab_example1}
\end{figure*}

\begin{figure*}[t]
    \centering
    \begin{minipage}[t]{0.48\textwidth}
        \centering
        \includegraphics[width=\linewidth]{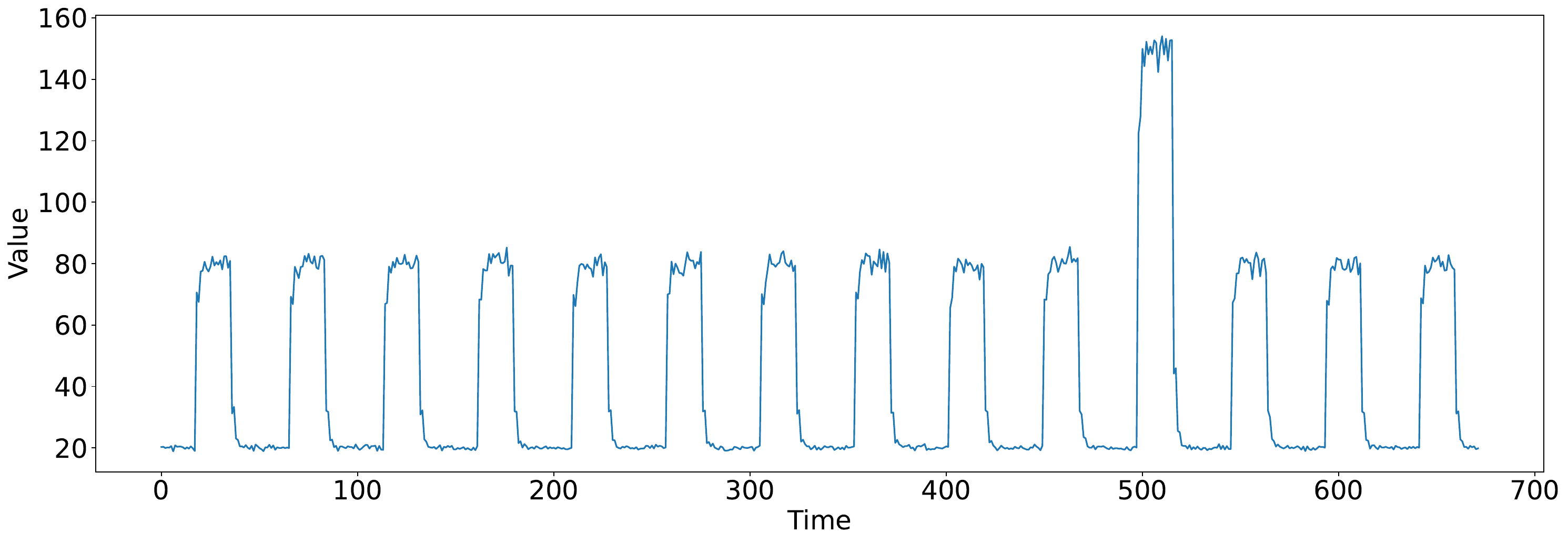}
        \subcaption{Input time series from the NAB dataset.}
    \end{minipage}%
    \hfill
    \begin{minipage}[t]{0.48\textwidth}
        \centering
        \includegraphics[width=\linewidth, height=4cm, keepaspectratio]{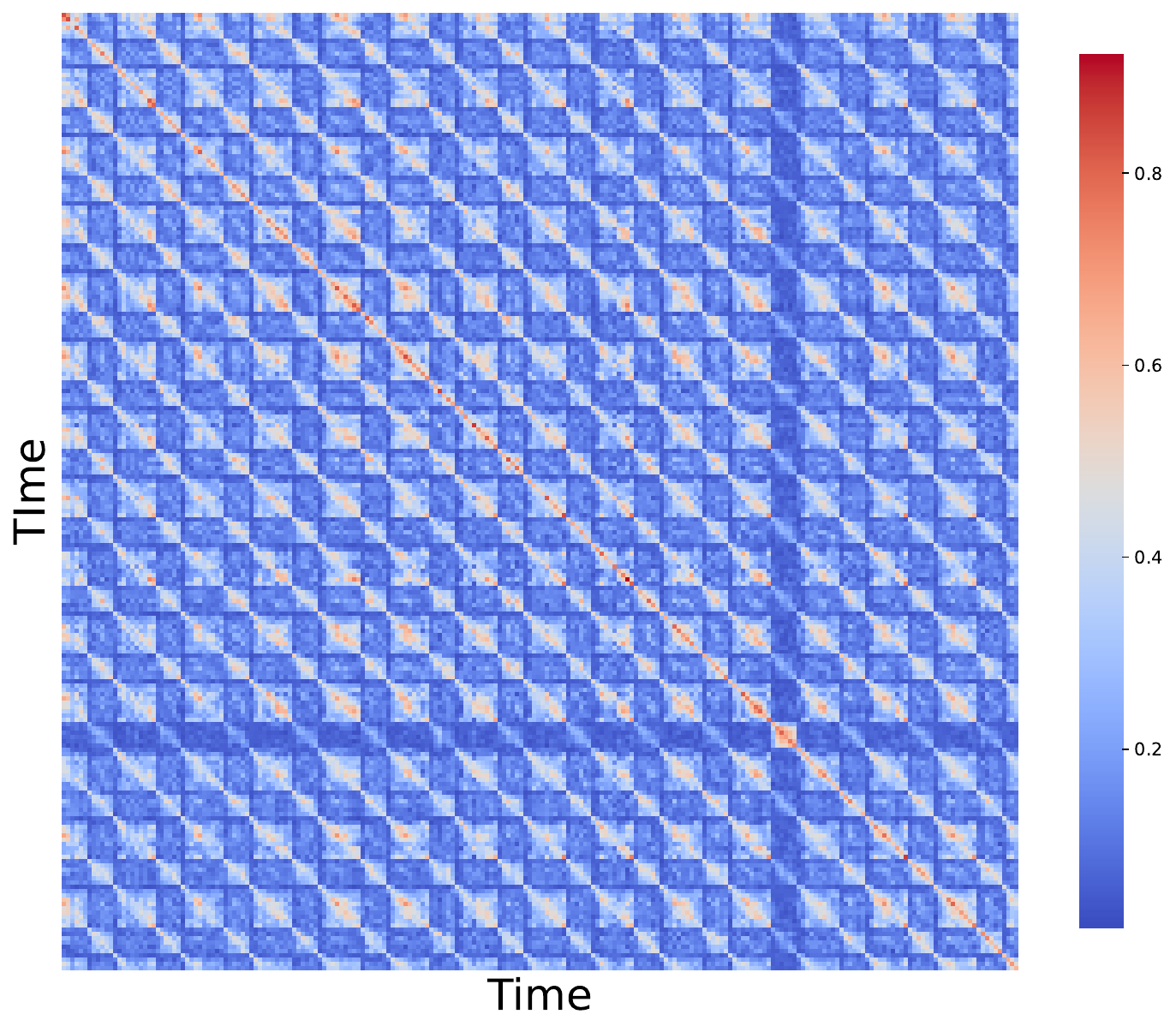}
        \subcaption{Heatmap of the similarity matrix $\mathbf{S}$. Anomalous points exhibit low similarity to normal regions.}
        \label{fig:example2_heatmap}
    \end{minipage}

    \vspace{0.5em}

    \begin{minipage}[t]{0.48\textwidth}
        \centering
        \includegraphics[width=\linewidth]{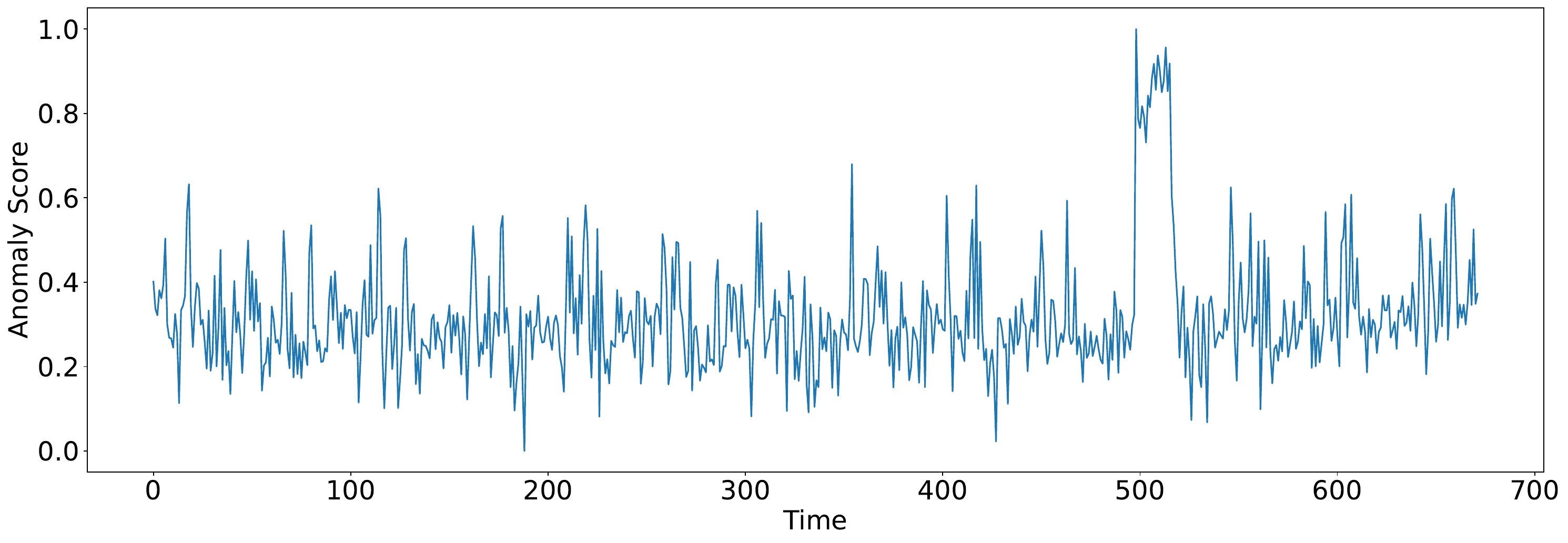}
        \subcaption{Anomaly scores from \textsc{Themis} using the spectral residual adapter.}
        \label{fig:example2_spectral}
    \end{minipage}%
    \hfill
    \begin{minipage}[t]{0.48\textwidth}
        \centering
        \includegraphics[width=\linewidth]{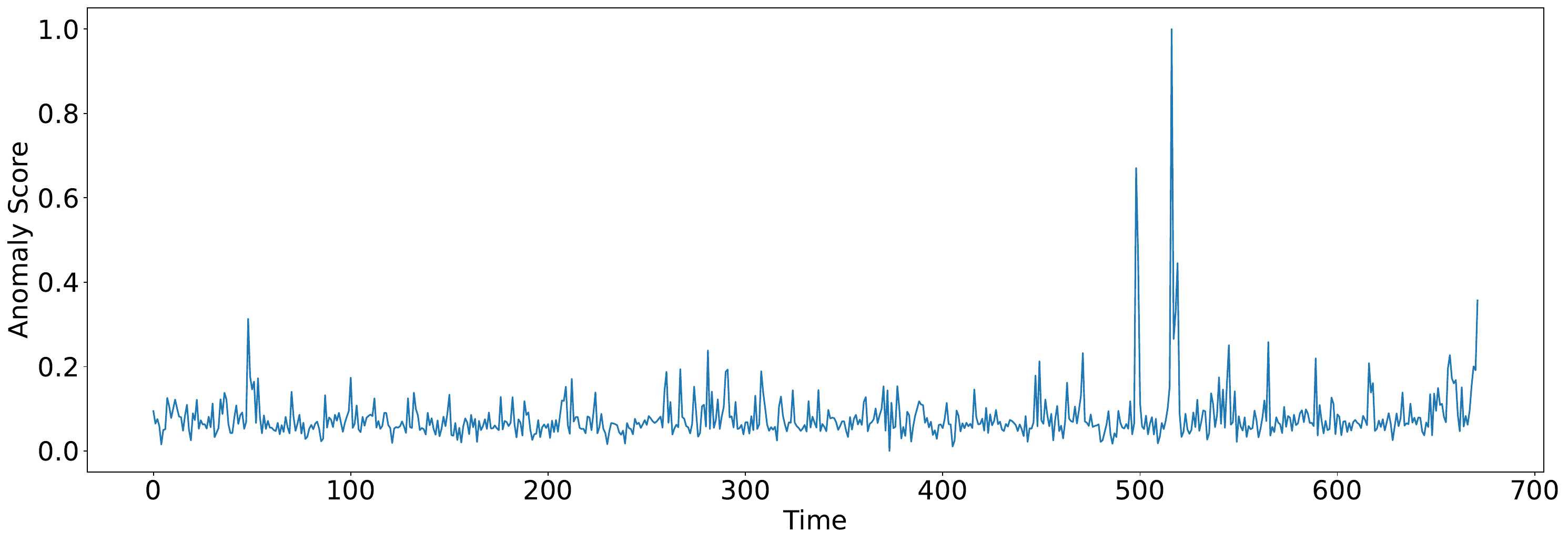}
        \subcaption{Anomaly scores from \textsc{Themis} using the LOF adapter.}
        \label{fig:example2_lof}
    \end{minipage}

    \caption{Example 1: Visual analysis on a time series from the NAB artificial anomaly dataset. \textbf{Top:} Input sequence (left) and corresponding similarity matrix (right), where anomalous points show reduced similarity to the rest of the series. \textbf{Bottom:} Anomaly scores computed by \textsc{Themis} using spectral residual (left) and LOF (right) adapters, both effectively highlighting anomalous regions.}
    \label{fig:nab_example2}
\end{figure*}

\begin{figure*}[ht]
    \centering
    \begin{minipage}[t]{0.48\textwidth}
        \centering
        \includegraphics[width=\linewidth]{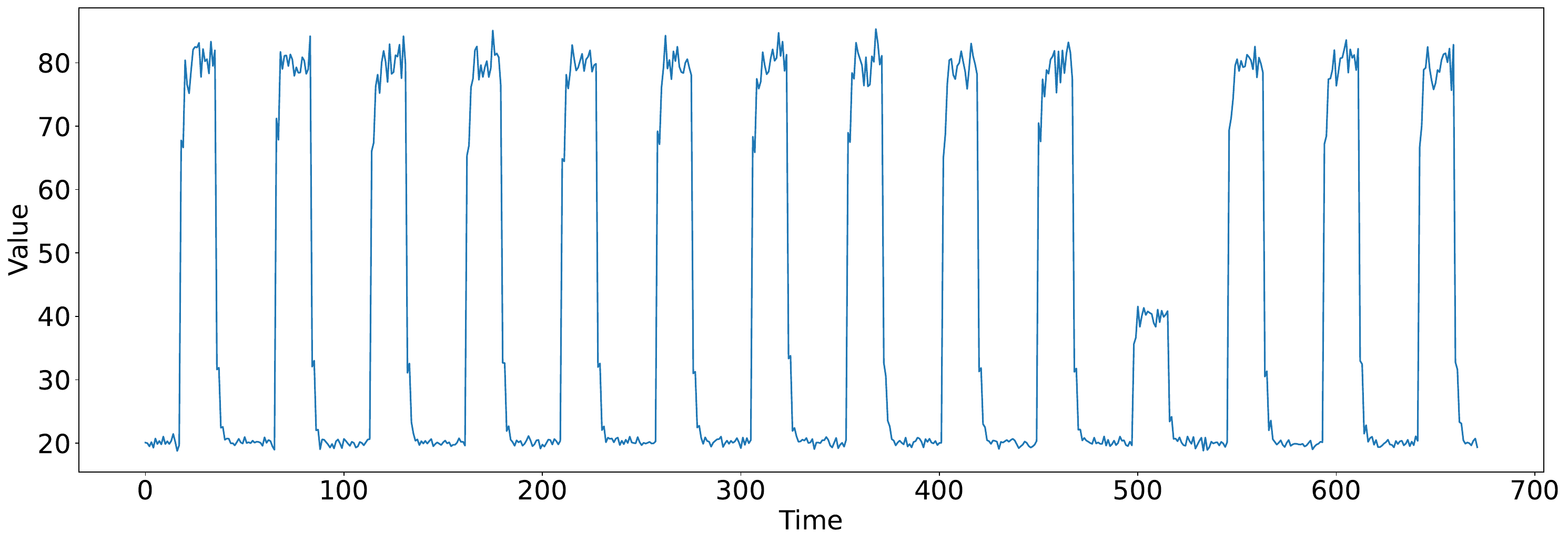}
        \subcaption{Input time series from the NAB dataset.}
    \end{minipage}%
    \hfill
    \begin{minipage}[t]{0.48\textwidth}
        \centering
        \includegraphics[width=\linewidth, height=4cm, keepaspectratio]{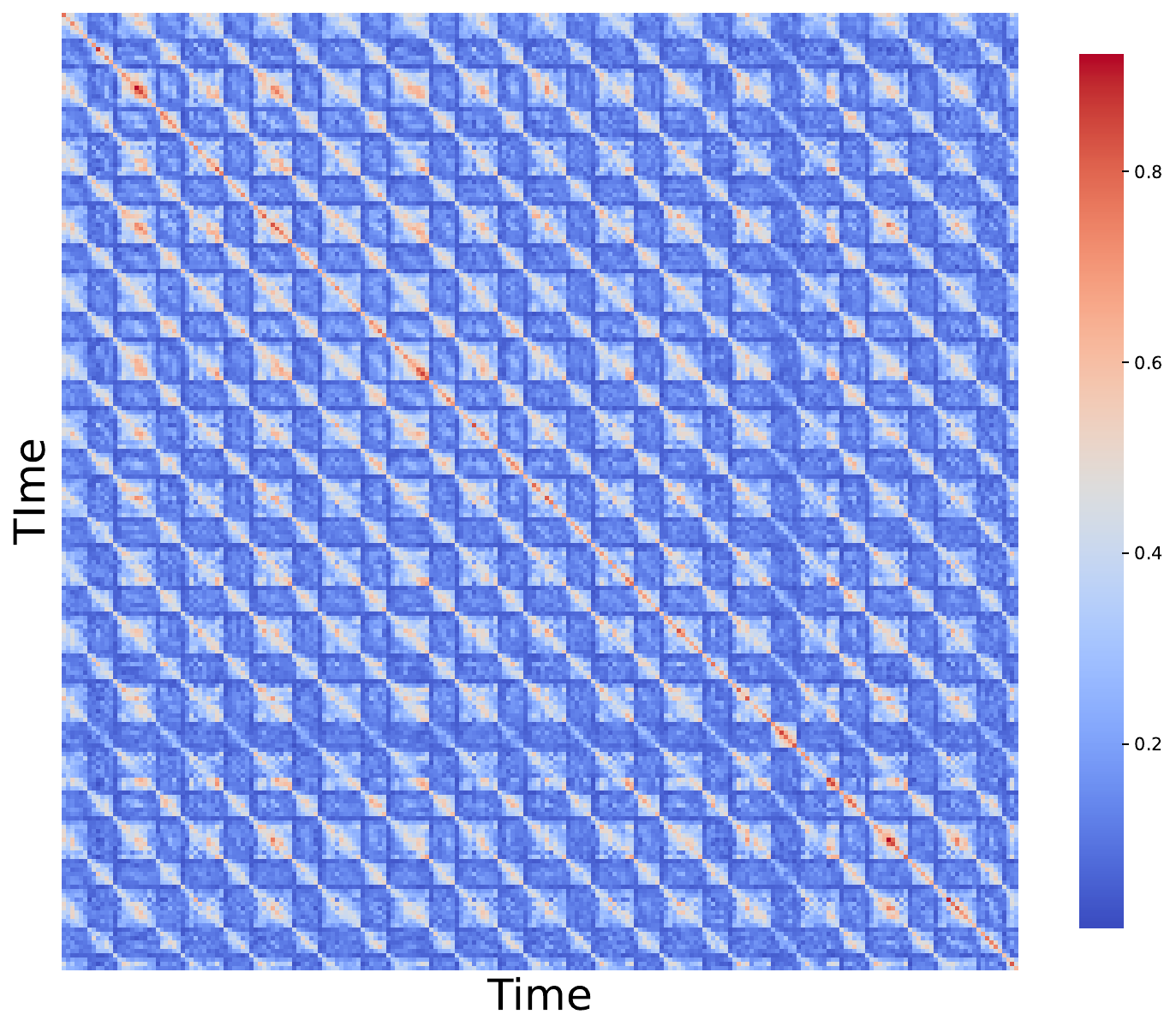}
        \subcaption{Heatmap of the similarity matrix $\mathbf{S}$. Anomalous points exhibit low similarity to normal regions.}
        \label{fig:example3_heatmap}
    \end{minipage}

    \vspace{0.5em}

    \begin{minipage}[t]{0.48\textwidth}
        \centering
        \includegraphics[width=\linewidth]{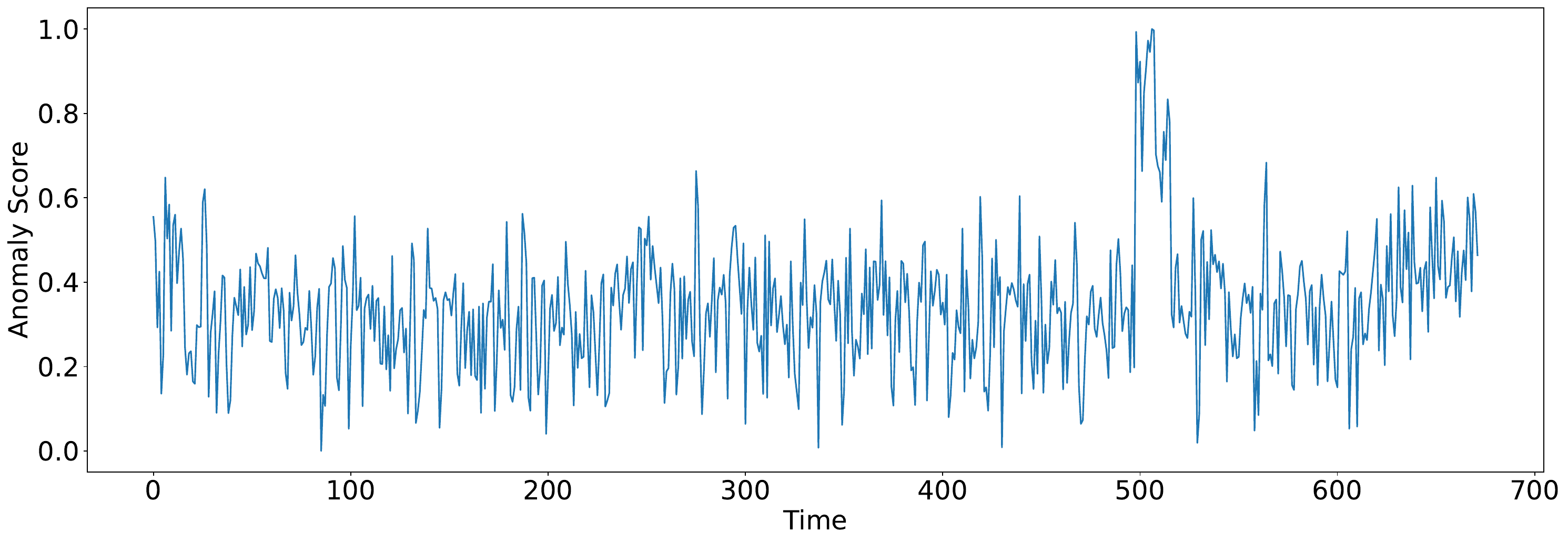}
        \subcaption{Anomaly scores from \textsc{Themis} using the spectral residual adapter.}
        \label{fig:example3_spectral}
    \end{minipage}%
    \hfill
    \begin{minipage}[t]{0.48\textwidth}
        \centering
        \includegraphics[width=\linewidth]{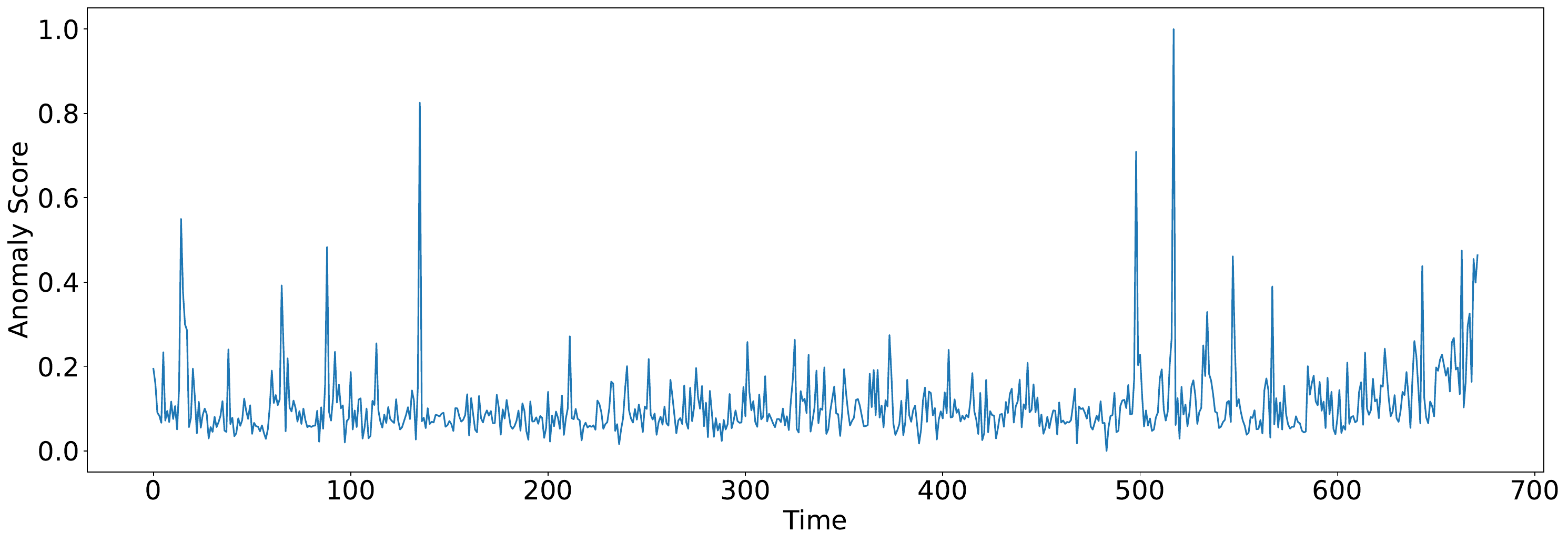}
        \subcaption{Anomaly scores from \textsc{Themis} using the LOF adapter.}
        \label{fig:example3_lof}
    \end{minipage}

    \caption{Example 2: Visual analysis on a time series from the NAB artificial anomaly dataset. \textbf{Top:} Input sequence (left) and corresponding similarity matrix (right), where anomalous points show reduced similarity to the rest of the series. \textbf{Bottom:} Anomaly scores computed by \textsc{Themis} using spectral residual (left) and LOF (right) adapters, both effectively highlighting anomalous regions.}
    \label{fig:nab_example3}
\end{figure*}

\begin{figure*}[ht]
    \centering
    \begin{minipage}[t]{0.48\textwidth}
        \centering
        \includegraphics[width=\linewidth]{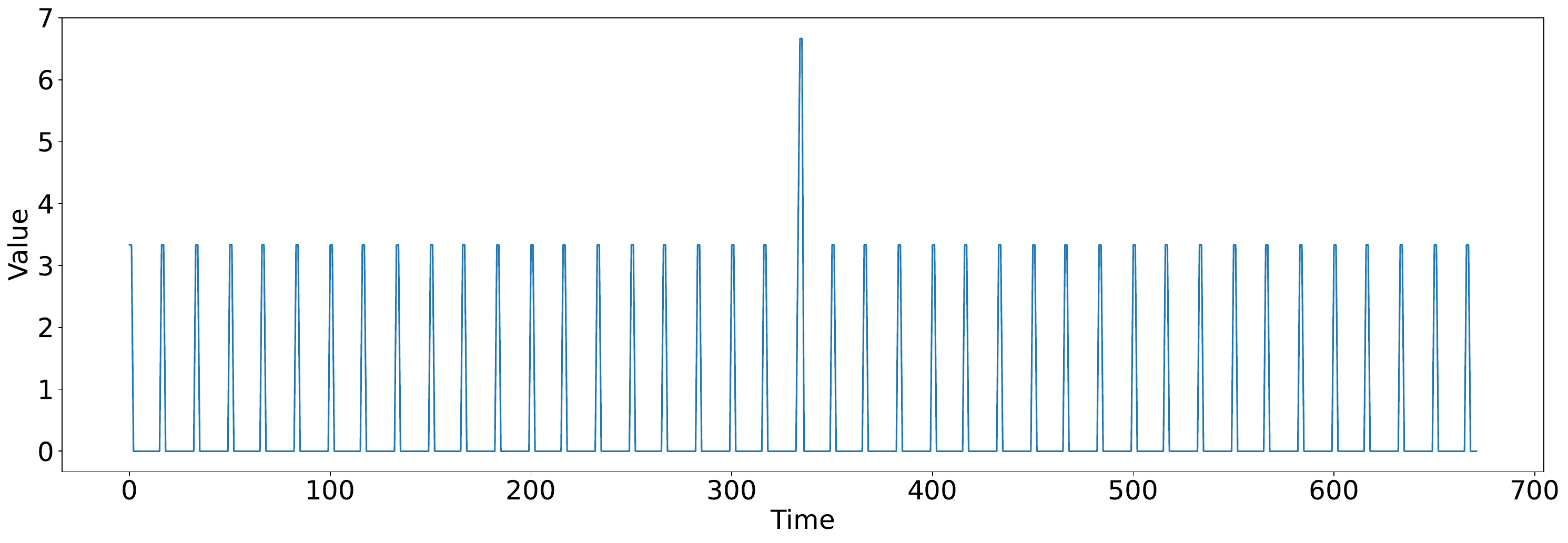}
        \subcaption{Input time series from the NAB dataset.}
    \end{minipage}%
    \hfill
    \begin{minipage}[t]{0.48\textwidth}
        \centering
        \includegraphics[width=\linewidth, height=4cm, keepaspectratio]{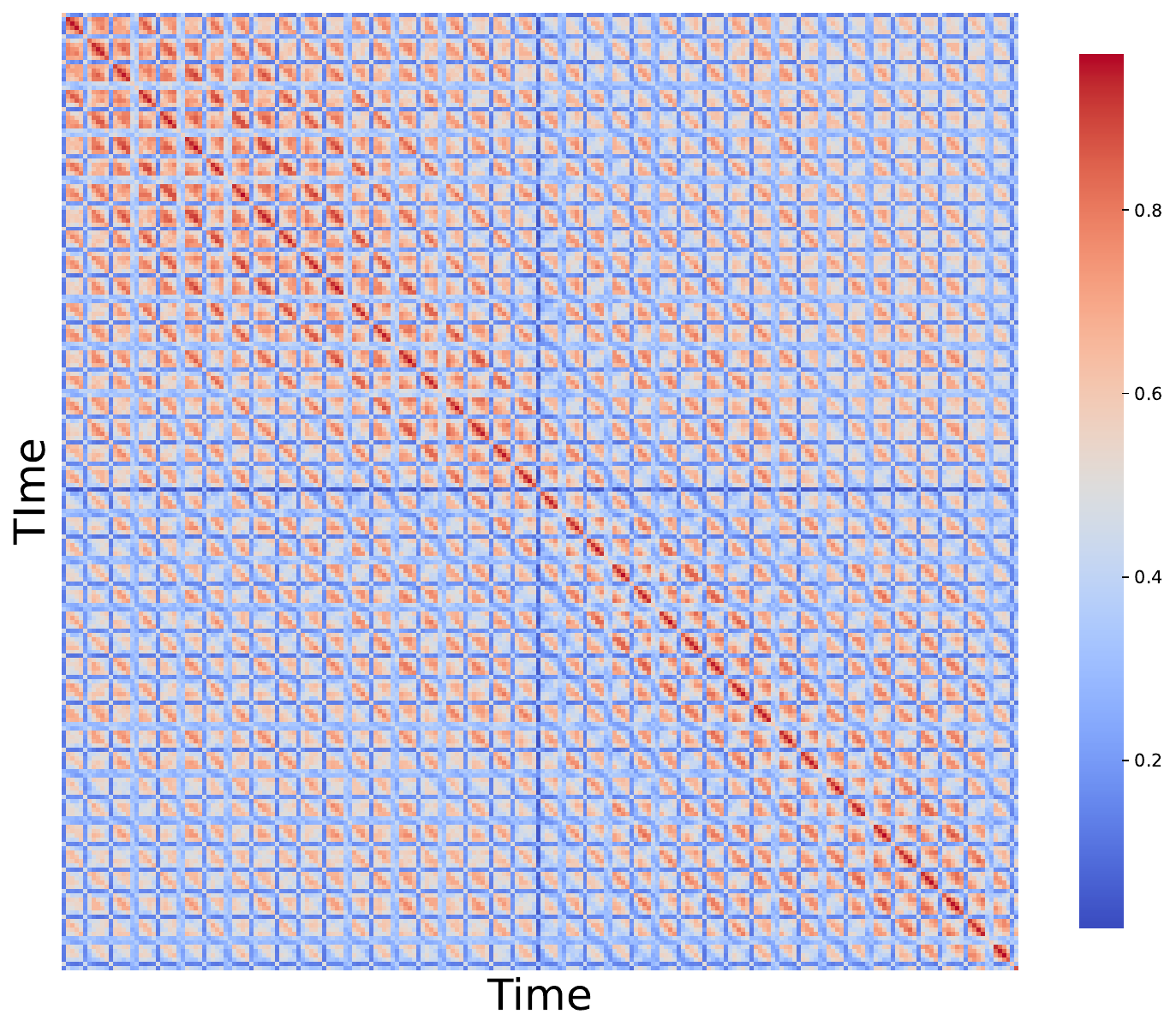}
        \subcaption{Heatmap of the similarity matrix $\mathbf{S}$. Anomalous points exhibit low similarity to normal regions.}
        \label{fig:example4_heatmap}
    \end{minipage}

    \vspace{0.5em}

    \begin{minipage}[t]{0.48\textwidth}
        \centering
        \includegraphics[width=\linewidth]{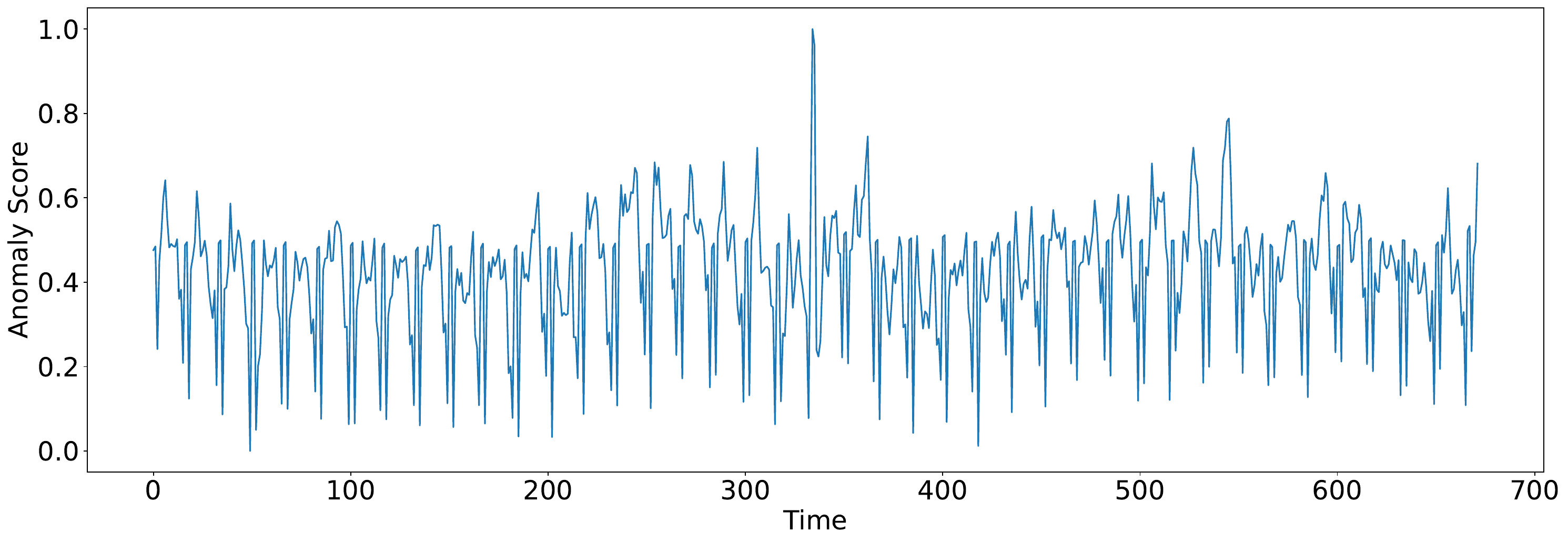}
        \subcaption{Anomaly scores from \textsc{Themis} using the spectral residual adapter.}
        \label{fig:example4_spectral}
    \end{minipage}%
    \hfill
    \begin{minipage}[t]{0.48\textwidth}
        \centering
        \includegraphics[width=\linewidth]{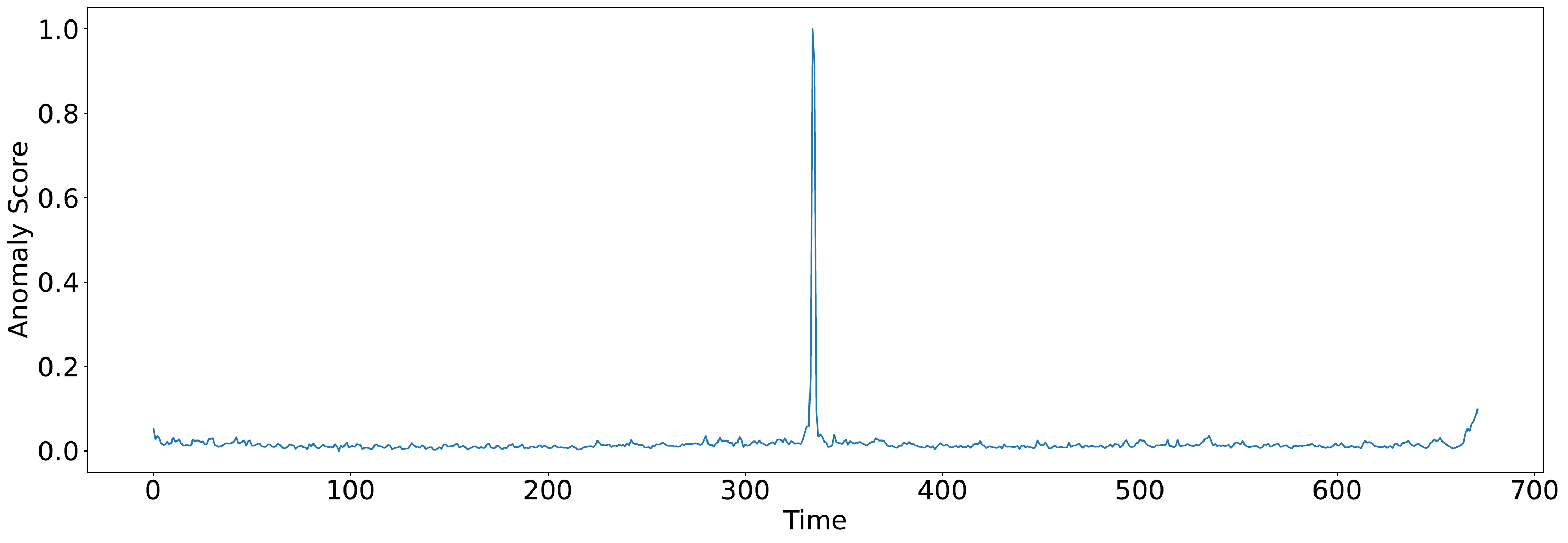}
        \subcaption{Anomaly scores from \textsc{Themis} using the LOF adapter.}
        \label{fig:example4_lof}
    \end{minipage}

    \caption{Example 4: Visual analysis on a time series from the NAB artificial anomaly dataset. \textbf{Top:} Input sequence (left) and corresponding similarity matrix (right), where anomalous points show reduced similarity to the rest of the series. \textbf{Bottom:} Anomaly scores computed by \textsc{Themis} using spectral residual (left) and LOF (right) adapters, both effectively highlighting anomalous regions.}
    \label{fig:nab_example4}
\end{figure*}

\section{Additional details of Anomaly Score Adapters}
\label{appendix:adapters}
\paragraph{(i) Spectral Residual Scoring}
Inspired by spectral graph analysis \cite{akoglu2015graph,ng2002spectral}, we perform an eigendecomposition of the WASM:

$$
\mathbf{S} = \mathbf{Q}\boldsymbol{\Lambda}\mathbf{Q}^\top,
$$

where $\boldsymbol{\Lambda} = \text{diag}(\lambda_1, \dots, \lambda_{B \cdot L})$ consists of eigenvalues sorted in ascending order, and $\mathbf{Q} = [\mathbf{q}_1, \dots, \mathbf{q}_{B \cdot L}]$ comprises the corresponding orthonormal eigenvectors. We form the spectral embedding $\mathbf{E} \in \mathbb{R}^{B \cdot L \times k}$ using the top-$k$ eigenvectors:

$$
\mathbf{E} = [\mathbf{q}_{B \cdot L - k + 1}, \dots, \mathbf{q}_{B \cdot L}].
$$

The anomaly score for each data point $t$ is computed using the complement of the normalized $\ell_2$-norm of its spectral embedding $\mathbf{e}_t$:

$$
s_t = 1 - \frac{\|\mathbf{e}_t\|_2}{\max_j \|\mathbf{e}_j\|_2}.
$$

Data points closely aligned with principal structures have higher spectral norms, indicating normality, while anomalous points exhibit lower norms.

\paragraph{(ii) Local Outlier Factor (LOF) Scoring}
To capture local density-based anomalies, we first convert the similarity matrix $\mathbf{S}$ into a corresponding distance matrix $\mathbf{D}$:

$$
D_{ij} = \max(\mathbf{S}) - S_{ij}.
$$

The LOF score for each data point $t$ is based on the local reachability density (LRD):

$$
\text{LRD}_k(t) = \left(\frac{1}{|N_k(t)|}\sum_{j \in N_k(t)} \max\{D_{tj}, \text{k-dist}(j)\}\right)^{-1},
$$

with $N_k(t)$ denoting the $k$-nearest neighbors. The LOF anomaly score quantifies local density disparity:

$$
\text{LOF}_k(t) = \frac{1}{|N_k(t)|}\sum_{j \in N_k(t)}\frac{\text{LRD}_k(j)}{\text{LRD}_k(t)}.
$$

Higher LOF values indicate anomalous points situated in lower-density neighborhoods.

\paragraph{(iii) Mean Similarity Scoring}
This method evaluates global contextual alignment by averaging similarity for each point $t$ with respect to all other points:

$$
\mu_t = \frac{1}{B \cdot L - 1}\sum_{j \neq t} S_{tj},
$$

and defines the anomaly score as the inverse of this average similarity:

$$
s_t = 1 - \mu_t.
$$

Lower average similarities indicate anomalous behavior.
\paragraph{(iv) Trimmed Top-$k$ Similarity Mean}
For enhanced robustness, we utilize a trimmed aggregation method. Letting $\mathbf{s}_t = \{ S_{tj} \mid j \neq t \}$, we sort and trim extreme fractions $\alpha$ from both ends:

$$
\mathbf{s}_t^{\text{trim}} = \text{sorted}(\mathbf{s}_t)[\lfloor \alpha n_t \rfloor : n_t - \lfloor \alpha n_t \rfloor],
$$

where $n_t = B \cdot L - 1$. The trimmed top-$k$ similarity mean is calculated as:

$$
t_t = \frac{1}{k}\sum_{j \in \text{TopK}(\mathbf{s}_t^{\text{trim}})} S_{tj},
$$

and the corresponding anomaly score is:

$$
s_t = 1 - t_t.
$$

Higher scores indicate reduced similarity to most contextually relevant neighbors.

\end{document}